\documentclass[10pt,journal,compsoc]{IEEEtran}

\ifCLASSOPTIONcompsoc
  \usepackage[nocompress]{cite}
  \usepackage[caption=false,font=footnotesize,labelfont=sf,textfont=sf]{subfig}
\else
  \usepackage{cite}
  \usepackage[caption=false,font=footnotesize]{subfig}
\fi
\usepackage{amsmath}
\usepackage{algorithmic}
\usepackage{url}
\usepackage{graphicx}
\usepackage{array}
\usepackage{amssymb}
\usepackage{multirow}
\usepackage[colorlinks]{hyperref}  

\usepackage{times}
\usepackage{epsfig}
 \usepackage{mathrsfs}
\usepackage{amsthm}
\usepackage{booktabs}
\usepackage{xcolor}
\usepackage[T1]{fontenc}
\usepackage{algorithmic}
\usepackage[ruled, vlined]{algorithm2e}

\SetKwInput{KwInput}{Input}             
\SetKwInput{KwOutput}{Output}

\usepackage{subfig}
\usepackage{colortbl}
\definecolor{mygray}{gray}{.9}

\theoremstyle{definition}
\newtheorem{definition}{Criterion}

\usepackage{mathtools}
\graphicspath{{fig/}}
\def\eg{\emph{e.g.}} 

\def\ie{\emph{i.e.}}

\def\etc{\emph{etc.}} 
\def\vs{\emph{vs. }}

\def\etal{\emph{et al. }}
\def\wrt{w.r.t. } 

\definecolor{cus_green}{HTML}{00B7A2}
\definecolor{cus_orange}{HTML}{FF9F8E}

\newcommand{\rev}[1]{{\color{black}{#1}}}
\newcommand{\majrev}[1]{{\color{black}{#1}}}

\begin{document}

\title{Combating Noisy Labels through Fostering Self- and Neighbor-Consistency}

\author{
Zeren~Sun,
Yazhou~Yao,
Tongliang Liu,
Zechao~Li,
Fumin~Shen,
and~Jinhui~Tang
\thanks{Corresponding author: Yazhou Yao.}
\IEEEcompsocitemizethanks{
	\IEEEcompsocthanksitem 
		Zeren Sun and Yazhou Yao are with the School of Computer Science and Engineering, Nanjing University of Science and Technology, Nanjing 210094, China, and also with the State Key Laboratory of Intelligent Manufacturing of Advanced Construction Machinery, Nanjing 210094, China. E-mail: \{zerens, ~yazhou.yao\}@njust.edu.cn.\protect
    \IEEEcompsocthanksitem 
        Tongliang Liu is with the School of Computer Science, Faculty of Engineering, the University of Sydney, Camperdown, NSW 2050, Australia. Email: tongliang.liu@sydney.edu.au\protect
	\IEEEcompsocthanksitem 
		Zechao Li and Jinhui Tang are with the School of Computer Science and Engineering, Nanjing University of Science and Technology, Nanjing 210094, China. E-mail: \{zechao.li, ~jinhuitang\}@njust.edu.cn.\protect	
    \IEEEcompsocthanksitem 
		Fumin Shen is with the School of Computer Science and Engineering, University of Electronic Science and Technology of China, Chengdu 610054, China. Email: fumin.shen@gmail.com.\protect
}
}

\markboth{}%
{Sun \MakeLowercase{\textit{et al.}}: Combating Noisy Labels through Fostering Self- and Neighbor-Consistency}

\IEEEtitleabstractindextext{%

\begin{abstract}

Label noise is pervasive in various real-world scenarios, posing challenges in supervised deep learning.
Deep networks are vulnerable to such label-corrupted samples due to the memorization effect.
One major stream of previous methods  concentrates on identifying clean data for training.
However, these methods often neglect imbalances in label noise across different mini-batches and devote insufficient attention to out-of-distribution noisy data. 
To this end, we propose a noise-robust method named Jo-SNC (\textbf{Jo}int sample selection and model regularization based on \textbf{S}elf- and \textbf{N}eighbor-\textbf{C}onsistency).
Specifically, we propose to employ the Jensen-Shannon divergence to measure the ``likelihood'' of a sample being clean or out-of-distribution.
This process factors in the nearest neighbors of each sample to reinforce the reliability of clean sample identification.
We design a self-adaptive, data-driven thresholding scheme to adjust per-class selection thresholds.
While clean samples undergo conventional training, detected in-distribution and out-of-distribution noisy samples are trained following partial label learning and negative learning, respectively.
Finally, we advance the model performance further by proposing a triplet consistency regularization that promotes self-prediction consistency, neighbor-prediction consistency, and feature consistency.
Extensive experiments on various benchmark datasets and comprehensive ablation studies demonstrate the effectiveness and superiority of our approach over existing state-of-the-art methods.
Our code and models have been made publicly available at \url{https://github.com/NUST-Machine-Intelligence-Laboratory/Jo-SNC}.
\end{abstract}


}

\maketitle

\IEEEdisplaynontitleabstractindextext

%
\IEEEpeerreviewmaketitle

\IEEEraisesectionheading{\section{Introduction} 
\label{sec:introduction}}  

\IEEEPARstart{T}{he} past decade has witnessed significant progress obtained by applying deep neural networks (DNNs) to various vision applications (\eg, image classification \cite{krizhevsky2012imagenet,szegedy2015}, object detection \cite{cai2024poly,cai2025cycle}, semantic segmentation \cite{pei2024cvpr,yao2021non}, \etc).
Although DNNs have demonstrated exceptional performance, their effectiveness is primarily due to the availability of large-scale, accurately annotated datasets such as ImageNet \cite{deng2009} and COCO \cite{lin2014microsoft}.
However, there are considerable challenges in amassing large-scale datasets with high-quality human annotations. This process can be laborious, time-consuming, and particularly demanding in areas where expertise is necessary for accurate labeling (\eg, fine-grained visual categorization \cite{xinyang2025} and medical image analysis \cite{wu2021meta}).
The hurdle of obtaining precise, large-scale annotations has emerged as a significant bottleneck for the practical application of deep neural networks.
Recently, one alternative solution has attracted increasing attention from researchers. 
This surrogate solution resorts to crowdsourcing platforms \cite{yan2014learning} or web image search engines \cite{fergus2010learning,schroff2010harvesting} to obtain low-quality data for model training \cite{liu2015classification,yao2018extracting,zhang2020web,zhang2020data,sun2020crssc,shu2023cmw}.
Unfortunately, while these datasets are typically more cost-effective and easier to acquire, they often yield noisy labels. Such labels can subsequently lead to model overfitting and negatively impact the performance of deep networks \cite{motivation2017,zhang2016understanding}.
Hence, it is imperative to address this predicament by developing robust learning methodologies capable of handling noisy labels.

Existing noise-robust methods predominantly fall into two categories: loss correction and sample selection.
Early approaches primarily attempt to correct losses during training. 
Some researchers propose to correct losses via the estimation of noise transition matrices \cite{patrini2017making,goldberger2017,hendrycks2018using}.
However, accurate estimation of this matrix entails prior knowledge or a subset of well-annotated data \cite{xia2020part,li2021provably,yang2022estimating}, which poses challenges in real-world applications.
Some methods focus on designing noise-robust loss functions to alleviate the overfitting of deep networks \cite{reed2015training,zhang2018generalized,tanaka2018joint}.
However, these methods may exhibit tendencies towards under-learning and are prone to fail in heavily-noisy cases.
Some approaches exploit label reassignment for loss correction \cite{tanaka2018joint,yi2019probabilistic,sun2021coldl,gong2022class}.
Despite potential benefits, these methodologies often necessitate complicated training processes and demonstrate unsatisfactory generalization performance when applied across diverse datasets.

\begin{figure*}[t]
\centering
\includegraphics[width=\linewidth]{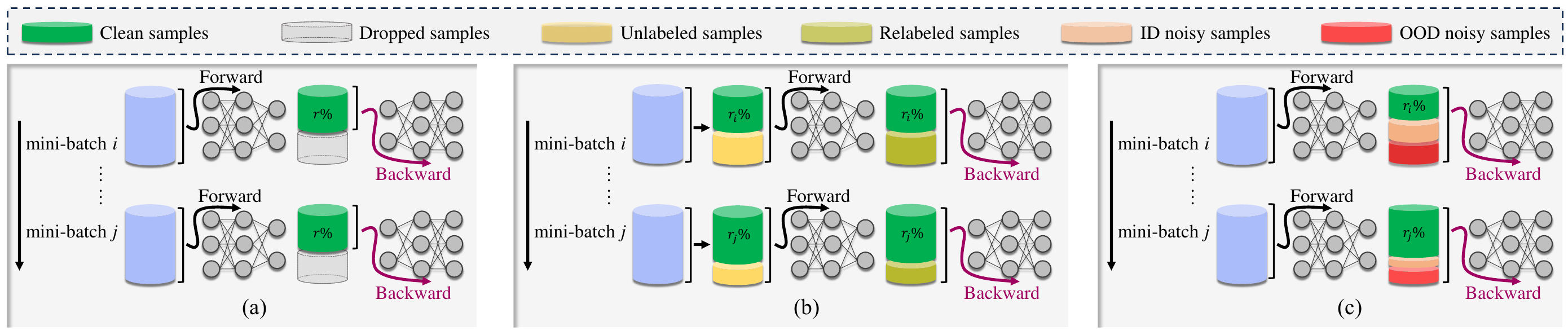}
\caption{
Comparative illustration of existing sample selection methods and our method.
\textbf{(a)} Early methods typically employ a predefined ratio to drop high-loss samples within each mini-batch.
\textbf{(b)} Recent approaches select clean samples from the entirety of training data via pre-epoch evaluation. The remaining samples are utilized in a semi-supervised manner.
\textbf{(c)} Our method distinguish clean, ID noisy, and OOD noisy samples within each mini-batch in an (approximately) global manner, and then leverage all training data accordingly.
}
\label{fig:motivation}
\end{figure*}


Sample selection is another promising strategy to counteract the detrimental effects of noisy labels. 
Methods following this paradigm aim to train deep networks using selected or reweighted training samples \cite{mentornet,ren2018learning,decoupling,coteaching,coteachingplus,wei2020combating,sun2020crssc}.
The primary challenge lies in establishing an appropriate criterion for identifying clean samples.
Recent studies have indicated that deep networks exhibit a memorization effect that makes them learn clean and simple patterns before overfitting noisy labels \cite{motivation2017,zhang2016understanding}. 
Accordingly, existing state-of-the-art (SOTA) methods (\eg, Co-teaching \cite{coteachingplus}, Co-teaching+ \cite{coteachingplus}, and JoCoR \cite{wei2020combating}) primarily advocate for the selection of small-loss samples as the clean ones through the use of a human-defined selection ratio. 
For example, Co-teaching \cite{coteaching} proposes to maintain two networks simultaneously and allow them to select small-loss samples for each other. 
Although notable performance gains have been witnessed by employing the small-loss sample selection strategy, these methods generally assume that noise ratios are consistent across all mini-batches. Hence, they perform sample selection within each mini-batch based on an estimated noise rate (as shown in Fig.~\ref{fig:motivation} (a)). However, this assumption may not hold true in real-world cases, and the noise rate is also challenging to estimate accurately (\eg, WebVision \cite{li2017webvision}).
Some recent researchers propose to replace the mini-batch-wise sample selection with global sample selection by identifying clean samples from the entirety of the training samples \cite{li2020dividemix,karim_unicon_2022,avidan_neighborhood_2022}. 
For example, Dividemix \cite{li2020dividemix} proposes to divide training data before each forward epoch by modeling all sample losses with a mixture model.
Despite their potential, these methods introduce a per-sample evaluation process on the entire training set before each training epoch to conduct global sample selection (as shown in Fig.~\ref{fig:motivation} (b)). This raises a computational efficiency problem, especially when learning with large-scale noisy datasets.

Moreover, a large body of prior literature primarily concentrates on closed-set scenarios, leaving open-set cases (\ie, real-world cases) comparatively under-explored \cite{sun2020crssc,xia2022extended}.
In the closed-set scenario, the true labels of noisy samples are exclusively derived from the known label space that is present within the training set, signifying that only in-distribution noisy samples occur.
On the contrary, in the open-set situation, the true labels of noisy samples are not guaranteed to be restricted to the known labels of the corresponding task. This implies the simultaneous existence of both in-distribution and out-of-distribution noisy samples. 
It is important to note that in open-set noisy datasets, high-loss samples are not necessarily associated with noisy labels. 
In fact, large loss values could be generated by hard samples, in-distribution (ID) noisy ones, and out-of-distribution (OOD) noisy ones. But the first two types of samples can potentially contribute to enhancing the robustness of deep networks \cite{sun2020crssc}.

To address the aforementioned issues, we propose an effective noise-robust approach named Jo-SNC (\textbf{Jo}int sample selection and model regularization based on \textbf{S}elf- and \textbf{N}eighbor-\textbf{C}onsistency).
Specifically, we first feed two different views of an image into a backbone network, allowing the network to predict two corresponding softmax probabilities accordingly. 
Then we divide samples globally based on two likelihood metrics in a high-efficiency manner.
We measure the likelihood of a sample being clean using the Jensen-Shannon (JS) divergence between its predicted probability distribution and its label distribution. 
Samples are deemed clean only if their likelihood of being clean or their nearest neighbors' likelihood of being clean is sufficiently high.
We measure the likelihood of a sample being out-of-distribution based on the prediction divergence between its two augmented views.
A sample is regarded as out-of-distribution noisy if the predictions derived from its two views diverge significantly.
By employing these two metrics, we enable (approximately) global sample selection within each mini-batch, thereby eliminating the requirement of per-sample evaluation before each epoch in existing global selection methods.
Moreover, we design a dynamic thresholding scheme, in which per-class thresholds are dynamically self-adjusted in a data-driven fashion, to promote adaptivity and class balance in sample selection.
Subsequently, clean samples undergo conventional training to fit their given labels.
In contrast, the labels of in-distribution and out-of-distribution noisy samples are reassigned by a mean-teacher model. 
For in-distribution samples, partial label learning is employed to refine their assigned pseudo-label distributions, which are then used during model training. 
For out-of-distribution instances, their complementary labels are exploited through negative learning to enhance model performance.
We further propose a triplet consistency regularization strategy comprising self-prediction consistency, neighbor-prediction consistency, and feature consistency. 
This triplet consistency regularization serves to bolster the separability between different types of label noise and further improve model performance. 
While feature consistency is fostered across the entire training set, the self- and neighbor-prediction consistency is only imposed on clean and in-distribution noisy samples. 
Consequently, our sample selection process and triplet consistency regularization are mutual-reinforced. 
Finally, our network is updated by jointly optimizing the classification loss and the triplet consistency regularization loss.
The comparison between typical sample selection methods and Jo-SNC is shown in Figure~\ref{fig:motivation}.
To summarize, our major contributions are as follows:

(1) We propose an effective noise-robust approach named Jo-SNC to alleviate the negative effect of noisy labels. 
Jo-SNC trains the network through a joint loss that incorporates a cross-entropy term alongside a triplet consistency term, thereby achieving enhanced classification and generalization performance.

(2) Jo-SNC facilitates the global selection of clean samples by utilizing the Jensen-Shannon divergence to evaluate the likelihood of each sample being clean. Both the samples themselves and their nearest neighbors are considered in determining cleanness. Additionally, we propose to distinguish between in-distribution noisy and out-of-distribution noisy samples based on the prediction divergence between different views of samples. Furthermore, we introduce a data-driven, self-adaptive thresholding scheme designed to dynamically adjust per-class selection thresholds.

(3) To fully exploit detected noisy data, beyond using a mean-teacher model to perform label correction, Jo-SNC proposes to employ partial label learning and negative learning to assist the network in learning from identified in-distribution noisy and out-of-distribution noisy samples, respectively. Moreover, we propose a triplet consistency regularization strategy, encompassing self-prediction consistency, neighbor-prediction consistency, and feature consistency, to further reinforce our sample selection process and advance model performance.

(4) We provide comprehensive experimental results on both synthetic and real-world noisy datasets to demonstrate the superiority of our proposed Jo-SNC compared with state-of-the-art methods. Extensive ablation studies are conducted to analyze the effectiveness of our approach.

This paper is an extended version of \cite{sun2021josrc}. The extensions include:
(1) assisting the global clean sample selection by additionally incorporating consistency from the nearest neighbors;
(2) improving the out-of-distribution detection through relaxing the identifying condition from prediction disagreement to prediction divergence;
(3) designing adaptive and class-specific selection thresholds based on temporal ensembling;
(4) employing partial-label-learning to refine pseudo-label distributions of selected in-distribution noisy labels;
(5) utilizing negative learning to \rev{exploit} complementary labels of detected out-of-distribution noisy samples;
(6) enriching consistency regularization from one level to three levels (\ie, self-prediction consistency, neighbor-prediction consistency, and feature consistency); 
(7) adding new state-of-the-art methods (\eg, UNICON \cite{karim_unicon_2022}, NCE \cite{avidan_neighborhood_2022}, SOP \cite{liu2022robust}, AGCE \cite{zhou2023asymmetric}, \etc) for comparison and re-evaluating quantitative experiments with our proposed Jo-SNC approach.


\section{Related Works}
\label{sec:related_works}
\subsection{Learning with noisy labels}

Existing literature on learning with noisy labels can be broadly classified into the following three categories \cite{sun2020crssc}: 1) loss correction methods, 2) sample selection methods, and 3) other methods.

\subsubsection{Loss correction methods}
A large proportion of existing studies on training with noisy labels focus on loss correction approaches. 
Some methods endeavor to estimate the noise transition matrix \cite{sukhbaatar2015iclr,chang2017active,patrini2017making,goldberger2017,hendrycks2018using}. 
For example, Patrini \etal \cite{patrini2017making} propose a loss correction method to estimate the noise transition matrix by using a deep network trained on the noisy dataset. 
Goldberger \etal \cite{goldberger2017} propose to employ an additional layer to model the noise transition matrix.
However, these methods are inherently limited as accurately estimating the noise transition matrix is challenging and may not be applicable in real-world scenarios. 
Some methods strive to design noise-tolerant loss functions \cite{reed2015training,zhang2018generalized}. 
For example, the bootstrapping loss \cite{reed2015training} extends the conventional cross-entropy loss with a perceptual term. 
Zhang \etal \cite{zhang2018generalized} propose a generalized cross-entropy loss by combining the mean absolute loss and the cross-entropy loss to enhance the robustness against noisy labels. 
However, these methods tend to underperform in real-world settings when the noise ratio is high. 
Some approaches devote to improving the quality of annotations by reassigning labels \cite{tanaka2018joint,yi2019probabilistic,sun2020crssc,sun2021coldl}.
For instance, Tanaka \etal \cite{tanaka2018joint} propose to alternatively update label distributions and network parameters. 
Sun \etal \cite{sun2021coldl} propose to train two networks simultaneously and let them learn label distributions for each other.
\majrev{NPN \cite{sheng2024adaptive} proposes to incorporate partial label learning (PLL) and negative learning (NL) to facilitate label estimation.
CA2C \cite{sheng2025ca2c} advances this design by introducing an asymmetric co-training framework that decouples PLL and NL, thereby achieving additional gains in label correction.}
Nevertheless, these methods tend to involve complicated training processes and do not consistently generalize well in diverse real-world scenarios.

\subsubsection{Sample selection methods}
An alternative strategy for dealing with noisy labels entails selecting and eliminating corrupted data. 
The challenge lies in finding suitable sample selection criteria. 
Previous studies have shown that DNNs tend to learn simple patterns first before memorizing noisy data \cite{motivation2017,zhang2016understanding}. 
Resorting to this observation, the small-loss sample selection criterion has been widely adopted: samples with lower loss values are more likely to be clean instances \cite{coteaching,coteachingplus,wei2020combating,sun2020crssc,li2020dividemix,sun2022boosting}. 
For example, Co-teaching \cite{coteaching} proposes to maintain two networks simultaneously during training, with one network learning from the other network’s selected small-loss samples.
JoCoR \cite{wei2020combating} proposes to use a joint loss, including the conventional cross-entropy loss and the co-regularization loss, to select small-loss samples.
However, the above methods select samples within each mini-batch based on a human-defined drop rate. 
In real-world scenarios, noise ratios in different mini-batches cannot be guaranteed identical, and the drop rate is challenging to estimate. 
\rev{Some researchers propose to leverage the grouping behavior of training data in the feature space to select clean samples.
For example, TopoFilter~\cite{neurips2020_topofilter} proposes to leverage the topological property of the data in feature space, identifying clean data by selecting the largest connected component of each class and dropping isolated data.
FINE~\cite{neurips2021_FINE} proposes to achieve clean sample selection by measuring the alignment between the latent distribution and each representation using the eigen decomposition of the gram matrix of the training data.
However, solely relying on the imperfect feature representations tends to result in suboptimal performance.
}
Some recent studies resort to some other selection criteria and conduct selection among all training samples \cite{li2020dividemix,karim_unicon_2022,avidan_neighborhood_2022}. 
For instance, Li \etal \cite{avidan_neighborhood_2022} propose to rely on contrastive neighbors in the feature space to determine whether a sample is noisy.
Karim \etal \cite{karim_unicon_2022} propose to divide samples before each epoch by using the JS divergence metric.
\majrev{ACT \cite{sheng2024enhancing} identifies different sample types by analyzing the agreement and discrepancy between models trained with asymmetric strategies. SplitNet \cite {kim2025splitnet} proposes a learnable module for
clean-noisy label splitting.}
However, these methods tend to necessitate evaluating the entire training set for performing sample selection before every epoch, which poses a computational efficiency problem in the training process, especially when dealing with large-scale datasets.

\subsubsection{Other methods for learning with noisy labels}
In addition to the aforementioned loss correction and sample selection methods, there are some other methods to cope with noisy labels \cite{peng2020suppressing,huang2020self,liu2020early,iscen2022learning,liu2022robust}.
For example, Peng \etal \cite{peng2020suppressing} propose to pay more attention to clean samples and less to noisy ones by grouping.
Huang \etal \cite{huang2020self} propose a self-adaptive training approach to address noisy labels by using the guidance from model predictions.
Liu \etal \cite{liu2020early} propose an early-learning regularization to prevent the network from memorizing noisy samples.
Nevertheless, these methods tend to yield sub-optimal performance due to the risk of under-learning.

\majrev{
\subsubsection{Learning with noisy labels in other domains}
Beyond image classification, learning with noisy labels has also been extended to other domains such as object detection~\cite{liu2022towards}, text classification~\cite{yuan2024hide}, re-identification~\cite{yang2024robust,yang2022learning}, and vision-language pretraining~\cite{huang2024noise}.
For instance, NLTE~\cite{liu2022towards} explores noise latent transferability to address noisy annotations for domain adaptive object detection.
Yuan \etal \cite{yuan2024hide} design a collaborative learning framework based on active learning to combine LLMs and small models for tackling noisy labels in text classification tasks.
LCNL~\cite{yang2024robust} addresses coupled noisy labels, which refers to noisy annotations and the accompanied noisy correspondence, in object re-identification.
Huang \etal \cite{huang2024noise} introduce NEVER to enhance vision-language pretraining robustness against noisy correspondence via positive and negative learning.
}

\begin{figure*}[t]
\centering
\includegraphics[width=\linewidth]{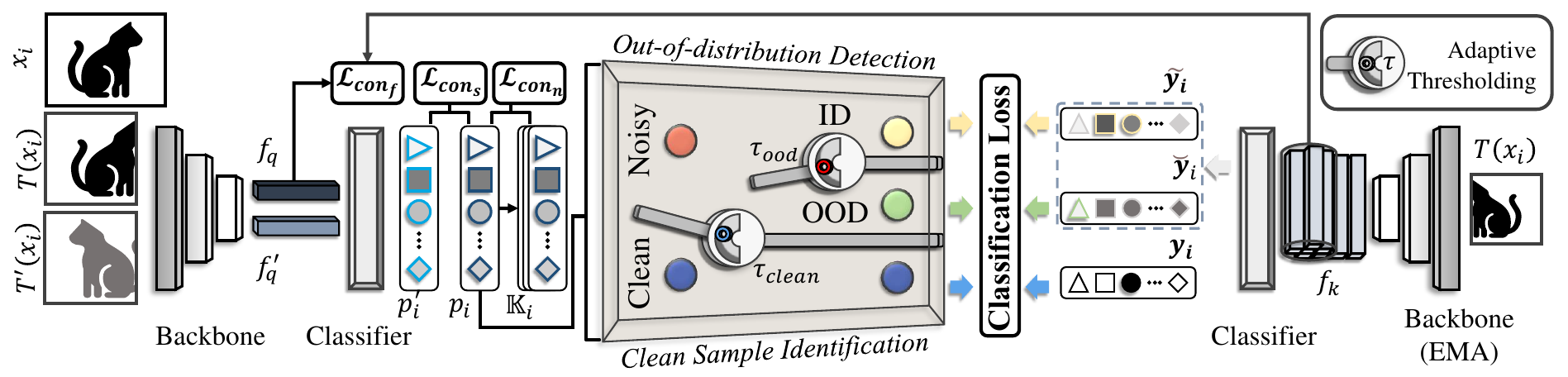}
\caption{Architecture of our proposed Jo-SNC.
Firstly, a sample selection strategy (comprising clean sample identification, out-of-distribution detection, and adaptive thresholding) is introduced to divide training data into three groups (\ie, clean, ID, and OOD).
Selected clean samples are leveraged for training conventionally. Conversely, identified ID and OOD noisy samples are learned by resorting to partial label learning and negative learning, respectively.
Moreover, a triplet consistency regularization is established, including self-prediction consistency $\mathcal{L}_{con_s}$, neighbor-prediction consistency $\mathcal{L}_{con_n}$, and feature consistency $\mathcal{L}_{con_f}$.
The final loss function is the integration of the classification loss and the consistency loss.
}
\label{fig:pipeline}
\end{figure*}

\subsection{Contrastive self-supervised learning}
Contrastive self-supervised learning aims to learn discriminative representations by using instance similarity \cite{simclr,he2020momentum,byol,simsiam}.
It advocates representation learning by driving representations of different views derived from the same image closer while distancing representations of different images \cite{sun2021coldl}.
SimCLR \cite{simclr} proposes to learn representations by maximizing agreement between augmented views of samples.
MoCo \cite{he2020momentum} proposes to facilitate self-supervised learning through a queue-based dictionary and a moving-averaged encoder.
BYOL \cite{byol} proposes to eliminate the demand for negative pairs by employing a momentum-encoder-based target network to bootstrap the outputs of the online network.
Chen and He \cite{simsiam} propose a contrastive self-supervised method with no need for negative sample pairs or momentum encoders.
In this work, we resort to MoCo-based contrastive self-supervised learning to achieve discriminative representations.

\subsection{Partial label learning}
Partial label learning (PLL) allows each training sample to be associated with a set of candidate labels, which invariably contain the ground-truth labels \cite{xu2021instance}.
Average-based methods are initial attempts to address this problem by treating all candidates equally \cite{hullermeier2006learning,cour2011learning,zhang2015solving}.
Nevertheless, such methods tend to yield inferior performance due to the misleading of false labels.
Identification-based methods regard ground-truth labels as latent variables and seek to identify them \cite{wang2019adaptive,xu2019partial,lyu2019gm,lv2020progressive}.
However, these methods retain a vulnerability to false labels due to the employment of hard labeling information.
Recently, soft-label-based methods have achieved promising performance in PLL problems \cite{wang2022pico}.
Inspired by these methods, this work proposes to leverage candidate label sets to promote the learning of identified noisy labels.

\subsection{Negative learning}
Negative learning (NL) is a type of indirect learning method for training neural networks \cite{ishida2017learning}.
It is proposed as a counterpart of conventional supervised learning (\ie, positive learning, PL).
While PL methods empower the model to predict ``which label the input sample belongs to'', NL methods teach the network ``the input sample does not belong to this complementary label''.
Since it is often easier to collect complementary labels than ground-truth ones, NL techniques can be seamlessly integrated into the positive learning process in various tasks. 
For example, \cite{gao2021discriminative} combines NL with PL in the classification task.
\cite{wang2022semi} employs unreliable pseudo-labels of pixels for semantic segmentation.
\cite{wei2022embarrassingly} proposes a semi-supervised method relying on negative learning.
When learning with noisy labels, complementary / false labels are more readily accessible than correct ones. 
Thus, this work resorts to negative learning to exploit selected out-of-distribution noisy samples.
We generate complementary labels for detected noisy samples and use them for NL, thereby improving model performance.


\section{The Proposed Method}
\label{sec:proposed_method}
In this study, we introduce a noise-robust method named Jo-SNC to alleviate the negative impact induced by noisy labels. 
The overview of our framework is illustrated in Fig.~\ref{fig:pipeline}.
The following sections begin with a brief introduction of preliminaries. We then elaborate on our proposed method in detail. Finally, we provide a succinct summary of our proposed Jo-SNC method.

\subsection{Preliminaries}

Generally, for a multi-class classification task with $C$ classes, we train deep neural networks using an accurately-labeled dataset $\mathbb{\hat{D}} = \{(x_i, \hat{y}_i)| 1 \le i \le N\}$, in which $x_i$ is the $i$-th training sample and $\hat{y}_i \in \{0, 1\}^C$ is its corresponding ground-truth one-hot label over $C$ classes. The conventional loss function is the cross-entropy between the predicted softmax probability distributions of training samples and their corresponding label distributions:
\begin{equation}\label{eq:celoss}
	\mathcal{L}_{CE} = - \frac{1}{N} \sum_{i=1}^{N} \sum_{c=1}^{C} \hat{y}_i^c \log(p^c_i),
\end{equation}
in which $ p^c_i $ is a simplified form of $p^c(x_i, \theta)$, denoting the predicted probability of sample $x_i$ for class $c$ given a model with parameters $\theta$.
However, for a dataset with noisy labels $\mathbb{D} = \{(x_i, y_i)| 1 \le i \le N\}$, the observed label $y_i$ of sample $x_i$ is not guaranteed to be consistent with its corresponding ground-truth label $\hat{y}_i$ (\ie, $y_i \neq \hat{y}_i$).
Thus, training networks using noisy datasets directly is problematic and usually leads to a dramatic performance drop, given the fact that DNNs have the capability to memorize all training samples, including noisy ones \cite{motivation2017}.

\textbf{Terminology}.
This paper adopts two consistency metrics to reveal how likely each sample could be clean or out-of-distribution (OOD). We accordingly term them as ``likelihood'', which is different from the traditional statistical concept of ``likelihood''.

\subsection{Consistency-driven sample selection}

\subsubsection{Clean sample identification}
Regarding samples with small cross-entropy losses as clean ones is one of the most widely-used sample selection criteria. 
This criterion is justified by the observation that DNNs tend to learn clean patterns first and then gradually fit noisy labels \cite{motivation2017,zhang2016understanding}. 
Methods using this criterion (\eg, Co-teaching \cite{coteaching} and Co-teaching+ \cite{coteachingplus}) typically select a pre-defined proportion of small-loss samples within each mini-batch.
Unfortunately, in real-world scenarios, noise ratios across different mini-batches inevitably fluctuate. 
Moreover, cross-entropy losses are unbounded, making it difficult to use a threshold instead of the ratio to select samples.
One potential solution is to record losses for all samples and select samples from the entire training set. 
However, this becomes impractical when the dataset volume is increasingly huge.

\begin{figure}[t]
\centering
\includegraphics[width=\linewidth]{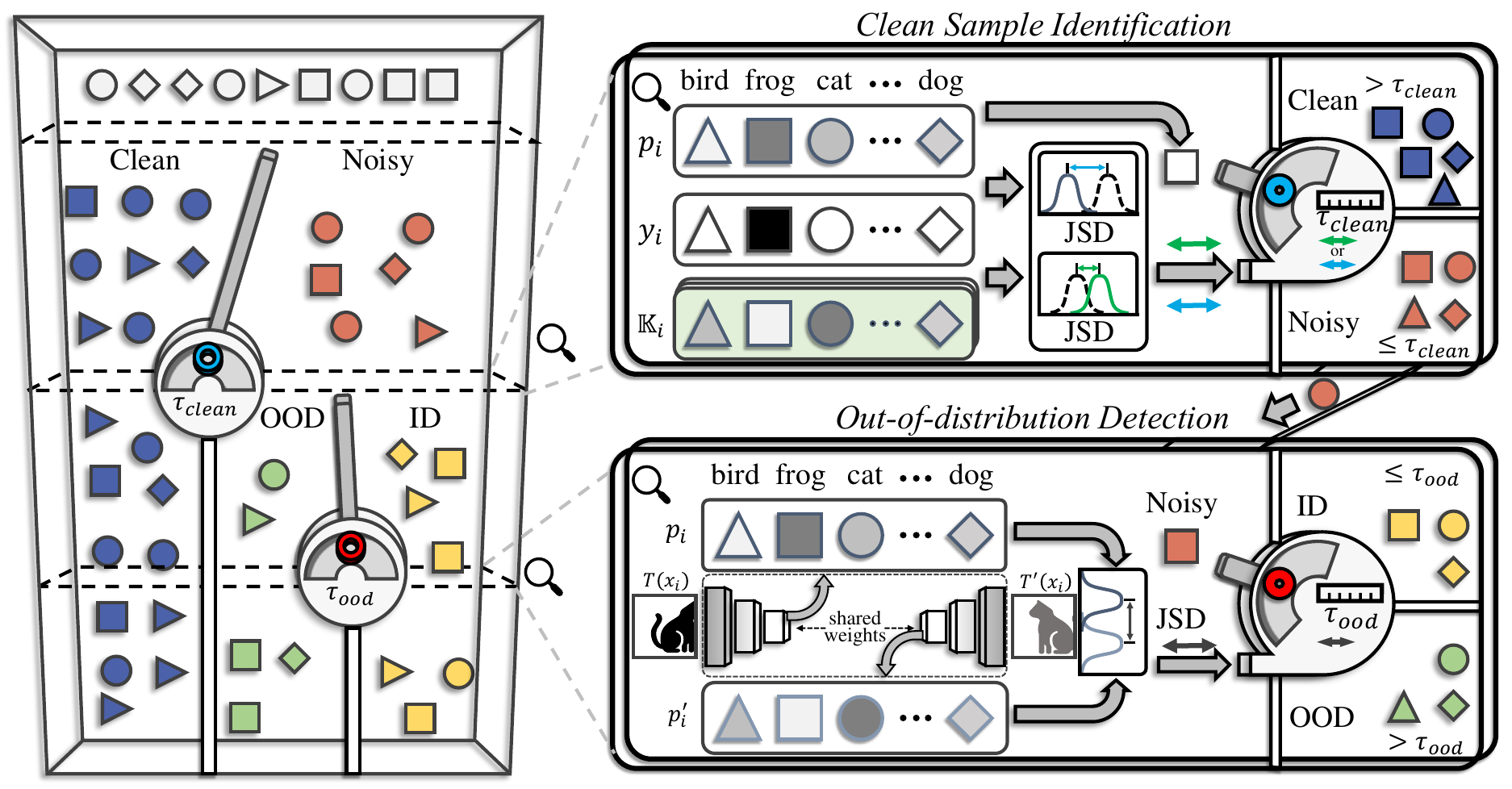}
\caption{The illustration of our sample selection strategy.
The proposed sample selection relies on two JS divergence-based metrics $\mathcal{P}_{clean}$ and $\mathcal{P}_{ood}$ to determine whether a sample is clean, ID, or OOD.
Moreover, we design a self-adaptive thresholding scheme to adjust selection thresholds $\tau_{clean}$ and $\tau_{ood}$ for each class in a data-driven manner.
}
\label{fig:pipeline_selection}
\end{figure}

To this end, we propose to reformulate the clean sample selection criterion from another perspective. Specifically, we propose to adopt the Jensen-Shannon (JS) divergence in Eq.~\eqref{eq:jsd_clean} to quantify the difference between the predicted probability distribution ${p_i} = [p_i^1, p_i^2, ..., p_i^C]$ and the observed label distribution ${y_i} = [y_i^1, y_i^2, ..., y_i^C]$ of the sample $x_i$ as follows:
\begin{equation}\label{eq:jsd_clean}
	\begin{split}
		d_i & = D_{JS}({p_i} \| {y_i}) \\
 		    & = \frac{1}{2} D_{KL} ({p_i} \| \frac{{p_i} + {y_i}}{2} ) + \frac{1}{2} D_{KL}({y_i} \| \frac{{p_i} + {y_i}}{2}),     
	\end{split}
\end{equation}
in which $D_{KL}(\cdot \| \cdot)$ is the Kullback-Leibler (KL) divergence function. 
The JS divergence is a measure of differences between two probability distributions. It is known to be bounded in $[0, 1]$, given a base 2 logarithm is used \cite{lin1991divergence}. 
Therefore, we can set a threshold to divide samples based on this metric to achieve efficient, approximately-global sample selection.
Intuitively, we can leverage $d_i$ to measure the ``likelihood'' of $x_i$ being clean as:
\begin{equation}\label{eq:prob_clean}
	\mathcal{P}_{clean}(x_i) = 1 - d_i \in [0, 1].
\end{equation}
In fact, $\mathcal{P}_{clean}(x_i)$ reveals the consistency between ${p_i}$ and ${y_i}$.
Here, we adopt smoothed label distributions \cite{szegedy2016} in calculating Eq.~\eqref{eq:jsd_clean} to avoid the issue of $\log(0)$.

Furthermore, we incorporate each sample with its nearest neighbors to reinforce the $\mathcal{P}_{clean}$-based sample selection. 
Specifically, for the sample $x_i$, we collect its $K$ nearest neighbors from feature space to form $\mathbb{K}_i = \{x_{i_j} \| 1 \le j \le K\}$.
We argue that if the $K$ nearest neighbors share the same observed label with $x_i$ and their average $\mathcal{P}_{clean}$ fall into the ``clean zone'', $x_i$ is also clean.
Finally, our clean sample selection criterion is defined as follows:
\begin{definition}\label{def:clean}
Given a sample $x_i \in \mathbb{D}$ whose observed label is $y_i$, it is identified as a clean sample if either one of the following two conditions is satisfied:
(1) $\mathcal{P}_{clean}(x_i) > \tau_{clean}$ (\ie, its ``likelihood'' of being clean is high enough);
(2) $\forall x_{i_j} \in \mathbb{K}_i, y_{i_j} = y_i$ and $\frac{1}{K} \sum_{j=1}^{K} \mathcal{P}_{clean}(x_{i_j}) > \tau_{clean}$ (\ie, its $K$ nearest neighbors share the same label with $x_i$ and their averaged ``likelihood'' of being clean is high enough).
\end{definition}

\textbf{Discussion.}
Similar to cross-entropy, the JS divergence serves as a measurement depicting differences between two probability distributions. 
Since the ${y_i}$ in Eq.~\eqref{eq:jsd_clean} remains unchanged in the back-propagation process, the JS divergence between ${p_i}$ and ${y_i}$ is equivalent to their cross-entropy. 
Accordingly, our proposed Criterion~\ref{def:clean} is consistent with the small-loss sample selection criterion. 
However, whereas the value of cross-entropy is not constrained, the JS divergence is bounded in $[0, 1]$, making it a natural global selection metric to describe how likely a sample could be clean.
By directly modeling the likelihood of a sample being clean using Eq.~\eqref{eq:prob_clean}, clean samples are selected more efficiently in an approximately-global manner with the threshold $\tau_{clean}$. 
This alleviates the issue caused by the imbalance of noise ratios within different mini-batches.
Moreover, by additionally incorporating the nearest neighbors, we reduce the negative effect of misleading predictions, thereby making the selection criterion less biased and improving the precision of clean sample selection.

\rev{Suffering from the unbounded value range of the cross-entropy loss, existing small-loss-based sample selection methods are either efficient but inaccurate (mini-batch-wise selection methods) or accurate but inefficient (global selection methods).
While mini-batch-wise sample selection methods are efficient during training (each training sample passes through the network only once per epoch), existing ones tend to yield unsatisfactory performance in real-world scenarios due to their implicit assumption of consistent noise ratios across mini-batches.
Contrarily, global sample selection methods, which evaluate each sample for selecting clean ones before every training epoch, perform better but incur significant computational costs (each training sample passes through the network at least twice per epoch), especially with large-scale noisy datasets.
Resorting to the characteristics of JS divergence discussed above, our method seeks to achieve a better trade-off between efficiency and performance, integrating the advantages of mini-batch-wise sample selection (each training sample passes through the network only once per epoch) and global sample selection (clean samples are selected within each mini-batch based on a global selection threshold per epoch).
By leveraging our JS-divergence-based sample selection criterion, our approach enables efficient, approximately global sample selection through mini-batch-wise identification instead of involving a per-sample evaluation process on the entire training set.
}

\subsubsection{Out-of-distribution detection}
Real-world scenarios contain both in-distribution (ID) noisy samples and out-of-distribution (OOD) noisy ones. 
Despite their noisy labels, they can contribute to the model if their labels are re-assigned properly, especially for ID samples.
Therefore, dropping all ``unclean'' samples directly is not data-efficient.

DNNs are usually uncertain about OOD samples when making predictions since their correct labels fall outside the task scope. 
Conversely, while ID noisy samples have corrupted labels, they usually lead to consistent model predictions.
Therefore, inspired by the self-supervised contrastive learning \cite{chen2020simple} and agreement maximization principle \cite{sindhwani2005co}, we propose to use the prediction consistency to distinguish ID and OOD noisy samples. 
Specifically, we first generate two augmented views $v_i = T(x_i)$ and $v'_i = T'(x_i)$ from a sample $x_i$ by applying two different image transformations $T(\cdot)$ and $T'(\cdot)$. These two views are subsequently fed into the network to produce their corresponding predictions ${p_i}$ and ${p'_i}$, respectively. 
Finally, we adopt the consistency between these two predictions to determine if this sample is out-of-distribution or not. 
More explicitly, we define the ``likelihood'' of a sample being out-of-distribution (OOD) as:
\begin{equation}\label{eq:prob_ood}
	\mathcal{P}_{ood}(x_i) = D_{JS}({p_i} \| {p'_i}).
\end{equation}
Consequently, given a proper $\tau_{ood} \in (0, 1)$, our OOD / ID sample selection criterion is defined as follows:

\begin{definition}\label{def:id_ood}
	Given a sample $x_i$ that is selected as a ``unclean'' one by Criterion~\ref{def:clean}, it is judged as an OOD noisy one if $\mathcal{P}_{ood}(x_i) > \tau_{ood}$ (\ie, its predictions of two differently augmented views diverge enough). If $\mathcal{P}_{ood}(x_i) \le \tau_{ood}$ (\ie, its predictions of two differently augmented views are consistent enough), it is deemed as an ID noisy sample.
\end{definition}

\textbf{Discussion.} 
Different from \cite{sun2021josrc}, we employ the JS divergence between predictions to determine whether a sample belongs to ID noise or OOD noise, rather than relying on a hard agreement / disagreement.
The disagreement-based OOD detection tends to suffer from false detection and error accumulation, especially in the later training stage where sample predictions tend to converge.
Our proposed JS divergence-based OOD detection is consistent with the disagreement-based OOD detection but relaxes its strict condition. 
Samples whose predictions of two different views are sufficiently divergent (even if the disagreement is not achieved) are also detected as OOD noise.
This mitigates the aforementioned limitation of disagreement-based detection.

\subsubsection{Adaptive selection thresholds}
To enable effective sample selection using Criterion~\ref{def:clean} and \ref{def:id_ood}, we need to properly design $\tau_{clean}$ and $\tau_{ood}$ and adjust them dynamically as training progresses. 
Moreover, since the difficulties in learning from different categories vary nontrivially, class-specific thresholds are more beneficial in distinguishing different types of samples from various categories.
Different from \cite{sun2021josrc} which adopts a unified $\tau_{clean}$ and gradually increases $\tau_{clean}$ in a linear manner using pre-defined initial and final values during training, we design a self-adaptive and class-specific threshold scheduling scheme.
Specifically, inspired by \cite{wang2023freematch}, we propose to generate thresholds for each class in a data-driven manner.
In each epoch, we record the $\mathcal{P}_{clean}$ and $\mathcal{P}_{ood}$ of each sample. 
We then calculate their mean values for each class $c$ as $\tilde{\tau}_{clean}(c) = \frac{1}{N_c} \sum_{x_i, y_i=c} \mathcal{P}_{clean}(x_i)$ and $\tilde{\tau}_{ood}(c) = \frac{1}{N_c} \sum_{x_i, y_i=c} \mathcal{P}_{ood}(x_i)$. $N_c$ denotes the number of samples in the class $c$.
Accordingly, we employ the $\tilde{\tau}_{clean}(c)$ and $\tilde{\tau}_{ood}(c)$ to assist the estimation of the thresholds for the next epoch.
It should be noted that the mean-based threshold estimation could be unstable between different epochs, which may lead to inferior stabilization of the model training.
Therefore, we employ temporal ensembling to refine the thresholds. 
The final threshold values for the class $c$ in the $i$-th epoch are defined as 
\begin{equation}\label{eq:threshold}
\left\{
    \begin{array}{l}
        \begin{aligned}  
        \tau_{clean}^{(t_i)}(c) =& \omega_{\tau} \tau_{clean}^{(t_i-1)}(c)  + (1 - \omega_{\tau}) \tilde{\tau}_{clean}(c),   \\
        \tau_{ood}^{(t_i)}(c)   =& \omega_{\tau} \tau_{ood}^{(t_i-1)}(c)  + (1 - \omega_{\tau}) \tilde{\tau}_{ood}(c). \\
        \end{aligned}\\
    \end{array}
\right.
\end{equation}
$\tau_{clean}^{(0)}(c)$ and $\tau_{ood}^{(0)}(c)$ are initialized as 0.
$\omega_{\tau}$ is set to a relatively small value (\eg, 0.75) in warmup epochs and a larger value (\eg, 0.975) afterward.
The intuition is that the threshold estimation is less reliable at the beginning and becomes increasingly stable as training progresses.
The moving-average mechanism effectively makes the training process more stable by alleviating the instability in the early stage of training.

Since models will get increasingly stronger during training and eventually overfit noisy labels, the value of $\tilde{\tau}_{clean}(c)$ increases as the training progresses.
Accordingly, our estimated threshold complies with a monotonically increasing design to ensure the effectiveness of noise identification. 
More samples will be treated as clean ones in initial epochs so that the model can learn simple and easy patterns from as many samples as possible. As the training proceeds, fewer samples are fed into the model as clean ones to ensure the quality of learned data.

\subsection{Robust learning on different samples}
The proposed Criterion~\ref{def:clean} and \ref{def:id_ood} jointly divide training data into three subsets: a clean subset $\mathbb{S}_{clean}$, an ID subset $\mathbb{S}_{id}$, and an OOD subset $\mathbb{S}_{ood}$. 
To leverage all training data effectively, we employ different learning strategies for different samples.

For samples in $\mathbb{S}_{clean}$, we keep their labels unaltered. To enhance the generalization performance, we adopt the label smoothing regularization (LSR) \cite{szegedy2016} when calculating their losses. Therefore, the label distribution of a clean sample $x_i$ is provided as Eq.~\eqref{eq:label_clean}, given its label $l_i \in \{1, 2, 3, ..., C \}$:
\begin{equation}\label{eq:label_clean}
	\tilde{y}_i^c = \left\{
		\begin{array}{ll}
			1 - \epsilon,			& {c = l_i} \\
			\frac{\epsilon}{C-1},	& {c \neq l_i}	
		\end{array},
	\right.
\end{equation}
in which $\epsilon$ controls the smoothness of the label distribution. 

For samples in ID subset $\mathbb{S}_{id}$, inspired by the mean-teacher model \cite{tarvainen2017mean}, we use the temporally averaged model (\ie, mean-teacher model) to generate pseudo labels.
While the top-1 prediction could be incorrect, we argue that the top-$\kappa$ predictions can provide more reliable supervision.
Therefore, we resort to partial label learning \cite{wang2022pico} to reformulate the pseudo-label distribution of ID samples.
Specifically, given an ID sample $x_i$, its partial label set $\Omega$ contains the top-$\kappa$ predicted labels derived from the mean-teacher model. 
The pseudo-label distribution of $x_i$ is provided as:
\begin{equation}\label{eq:label_id}
	\tilde{y}_i = \sigma(\frac{p(x_i, \theta_{mt})}{T_{pll}}),
\end{equation}
in which $\sigma(\cdot)$ is the softmax function. $\theta_{mt}$ denotes parameters of the mean-teacher model.
$T_{pll}$ is a temperature factor used to emphasize predictions on labels in the partial label set $\Omega$. 
In our implementation, we empirically set $T_{pll}=0.1$ when $c \in \Omega$, and $T_{pll}=1$ when $c \notin \Omega$.
Accordingly, $\tilde{y}_i^c$ will be sufficiently intensified when $c \in \Omega$, guiding the prediction to comply with the partial label set $\Omega$ during training.
The classification loss for samples in $\mathbb{S}_{clean} \cup \mathbb{S}_{id}$ are computed by
\begin{equation}\label{eq:pl_loss}
    \mathcal{L}_{ce}(x_i) = - \sum_{c=1}^{C} \tilde{y}_i^c \log(p^c_i).
\end{equation}

While positive learning is performed on samples in $\mathbb{S}_{clean} \cup \mathbb{S}_{id}$, we propose to learn OOD noisy samples differently.
Since the true labels of OOD samples fall outside the task scope, the DNN will be confused when predicting their label assignments and cannot correctly find their ground-truth labels.
However, although correct labels are barely accessible for OOD samples, it is considerably easier to find labels that OOD samples do not belong to \cite{kim2019nlnl,wei2022embarrassingly}.
Therefore, we propose to exploit OOD samples by negative learning to boost generalization performance.
In practice, given an OOD sample $x_i \in \mathbb{S}_{ood}$, we obtain its negative label $\breve{y}_i^c$ by using the lowest predictions from the mean-teacher model as:
\begin{equation}\label{eq:nl_label}
\breve{y}_i^c = \left\{
		\begin{array}{ll}
			1,	& {c = \arg\min_{c} p^c(x_i, \theta_{mt}) } \\
			0,	& {otherwise}	
		\end{array}.
	\right.
\end{equation}
Then, the negative learning loss is calculated as 
\begin{equation}\label{eq:nl_loss}
    \mathcal{L}_{nl}(x_i) = - \sum_{c=1}^C \breve{y}_i^c \log(1 - p_i^c).
\end{equation}

It should be noted that the mean-teacher model is not updated via the loss back-propagation. Instead, its parameters  $\theta_{mt}$ is an exponential moving average of $\theta$. Specifically, given a decay rate $\omega \in [0, 1]$, $\theta_{mt}$ is updated in each training step as follows:
\begin{equation}\label{eq:mt_update}
	\theta_{mt} \leftarrow \omega \theta_{mt} + (1- \omega) \theta.
\end{equation}
In our implementation, $\omega$ is empirically set to 0.99.

\begin{algorithm}[t]\small
	\label{alg}
	\SetAlgoLined
	\KwInput{Network $\theta$, mean-teacher $\theta_{mt}$, learning rate $\eta$, iteration $I_{\max}$, epoch $t_w$ and $t_{\max}$.}
    \For{$t = 1, ..., t_{w}$}
    {
        \For{$\text{iter} = 1, 2, 3, ..., I_{\max}$}
		{
            Sample a mini-batch $\mathbb{B}$ randomly. \\
            Compute $\mathcal{L}_{CE}$ using entire $\mathbb{B}$ by Eq.~\eqref{eq:celoss}.\\
            Update $\theta \leftarrow \theta - \eta \nabla \mathcal{L}_{CE} $.\\
            Update $\theta_{mt}$ by Eq.~\eqref{eq:mt_update}.\\
            Update the queue $\mathscr{Q}$. \\
        }
        Update $\tau_{clean}$ and $\tau_{ood}$ according to Eq.~\eqref{eq:threshold}
    }
	\For{$t = t_{w}, ..., t_{\max}$}
	{
		\For{$\text{iter} = 1, 2, 3, ..., I_{\max}$}
		{
			Sample a mini-batch $\mathbb{B}$ randomly. \\
			Divide samples into $\mathbb{B}_{clean}$, $\mathbb{B}_{id}$, and $\mathbb{B}_{ood}$ based on Criteria~\ref{def:clean} and \ref{def:id_ood}.\\
            Obtain $\tilde{y}$ for $x \in \mathbb{B}_{clean}$ by Eq.~\eqref{eq:label_clean}.\\
            Obtain $\Omega$ and refine $\tilde{y}$ for $x \in \mathbb{B}_{id}$ by Eq.~\eqref{eq:label_id}. \\
            Obtain negative labels $\breve{y}$ for $x \in \mathbb{B}_{ood}$ by Eq.~\eqref{eq:nl_label}. \\
            Compute $\mathcal{L}_{cls}$ according to Eq.~\eqref{eq:loss_cls}. \\
            Compute $\mathcal{L}_{con_s}$, $\mathcal{L}_{con_n}$, and $\mathcal{L}_{con_f}$ according to Eqs.~\eqref{eq:loss_con_pred}, \eqref{eq:loss_con_neighbor}, and \eqref{eq:loss_con_feat}.\\
            Compute $\mathcal{L}$ using entire $\mathbb{B}$ according to Eq.~\eqref{eq:loss_final}.\\
            Update $\theta \leftarrow \theta - \eta \nabla \mathcal{L}$.\\	
			Update $\theta_{mt}$ by Eq.~\eqref{eq:mt_update}.\\
            Update the queue $\mathscr{Q}$. \\
		}
        Update $\tau_{clean}$ and $\tau_{ood}$ according to Eq.~\eqref{eq:threshold}
	}
	\KwOutput{Updated network $\theta$.} 
	\caption{Jo-SNC}
\end{algorithm}

\subsection{Triplet consistency regularization}

As stated above, we use the prediction consistency of each sample to measure its likelihood of being OOD. 
We follow the intuition that in-distribution samples (including clean ones and noisy ones) tend to produce consistent predictions while out-of-distribution samples do not.
Thus, we propose to use a self-prediction consistency loss as Eq.~\eqref{eq:loss_con_pred} to provide joint supervision 
for enhancing the separability between ID and OOD samples.
\begin{equation}\label{eq:loss_con_pred}
	\mathcal{L}_{con_s} = \frac{1}{N} \sum_{i=1}^N \rho_i (D_{KL}({p_i} \| {p'_i}) + D_{KL}({p'_i} \| {p_i})),
\end{equation}
in which $\rho_i = 1$ if $x_i \in \mathbb{S}_{clean} \cup \mathbb{S}_{id}$; otherwise, $\rho_i = 0$.
Resorting to this self-prediction consistency regularization term, clean samples and ID noisy samples are encouraged to make consistent predictions.
Our approach is accordingly able to select clean / ID / OOD samples more effectively.

In addition to self-prediction consistency, inspired by \cite{iscen2022learning}, we further promote neighbor-prediction consistency by using a neighbor-consistency regularization loss
\begin{equation}\label{eq:loss_con_neighbor}
    \mathcal{L}_{con_n} = \frac{1}{N} \sum_{i=1}^N \rho_i D_{KL}({p_i}\| \sum_{x_{i_j}, x_{i_{j^{\prime}}} \in \mathbb{K}_i} \frac{s_{i,i_j}}{\sum_{i_{j^{\prime}}} s_{i, i_{j^{\prime}}}} {p_{i_j}}).
\end{equation}
$s_{i, i_{j}}$ denotes the cosine similarity between $x_i$ and its neighbor $x_{i_j}$.

The Criterion~\ref{def:clean} and the Eq.~\eqref{eq:loss_con_neighbor} both rely on the nearest neighbors, which demands high-quality extracted features. 
To enhance feature extraction, we resort to feature consistency encouraged by self-supervised learning.
Specifically, we follow MoCo \cite{he2020momentum} and treat the network and its mean-teacher model as the query network and key network. 
A multi-layer perception layer is added to project the last-convolutional-layer feature of sample $x$ to the query embedding $f_q$ and key embedding $f_k$ (both L2-normalized). 
A queue $\mathscr{Q}$ is maintained to store the most current key embeddings and is updated chronologically.
The feature consistency regularization loss is thus formulated as 
\begin{equation}\label{eq:loss_con_feat}
    \mathcal{L}_{con_f} = - \frac{1}{N} \sum_{x \in \mathbb{D}} \log \frac{exp(f_q^{\top} \cdot f_{k+}/T_{ssl})}{\sum_{f_k^{\prime} \in \mathscr{P}} exp(f_q^{\top} \cdot f_k^{\prime}/T_{ssl})},
\end{equation}
in which $f_{k+}$ \rev{represents the positive key embedding} and $\mathscr{P} = \{f_k\} \cup \mathscr{Q}$ denotes the embedding pool. For simplicity, $T_{ssl}$ is empirically set to 0.1 by default.

\textbf{Discussion}.
Our method incorporates a three-level consistency regularization to boost model performance. 
First, self-prediction consistency promotes consistent predictions among in-distribution samples (including clean and noisy ones). Accordingly, the prediction consistency-based OOD detection is achieved more effectively.
Secondly, neighbor-prediction consistency encourages in-distribution samples to share consistent predictions with their nearest neighbors. This implicitly benefits the neighbor-based selection as defined in Criterion~\ref{def:clean}.
Lastly, feature consistency promotes the representation and compactness of extracted features. Better feature extraction is favorable for the acquisition of more reliable nearest neighbors within the feature space. Moreover, we use the maintained queue $\mathscr{Q}$ to find the nearest neighbors for each sample, thereby enhancing memory usage and computational efficiency.

\subsection{Summary}
Integrating all submodules, our final objective loss function is 
\begin{equation}\label{eq:loss_final}
	\mathcal{L} = \mathcal{L}_{cls} + \alpha \mathcal{L}_{con_s} + \beta \mathcal{L}_{con_n}  + \gamma \mathcal{L}_{con_f}.
\end{equation}
$\alpha$, $\beta$, and $\gamma$ are hyper-parameters to balance different loss terms.
$\mathcal{L}_{cls} = \frac{1}{N} \sum_{i=1}^N \mathcal{L}_{cls}(x_i)$, in which 
\begin{equation}\label{eq:loss_cls}
	\mathcal{L}_{cls}(x_i) = \left\{
		\begin{array}{ll}
			\mathcal{L}_{ce}(x_i),	& {x_i \in \mathbb{S}_{clean} \cup \mathbb{S}_{id}} \\
			\mathcal{L}_{nl}(x_i),	& {x_i \in \mathbb{S}_{ood}}	
		\end{array}.
	\right.
\end{equation}

In summary, Jo-SNC abides by a selection-correction training paradigm.
Firstly, samples are divided into three subsets (\ie, clean subset, ID noisy subset, and OOD noisy subset).
Afterward, different learning strategies are applied depending on the noise type of samples.
During this learning process, we take the imperfection of the sample selection into account and propose to employ label smoothing regularization, partial label learning, and negative learning accordingly to improve error tolerance and robustness.
Lastly, we impose triplet consistency regularization to further boost the performance of our proposed model.
Details of our method are shown in Algorithm~\ref{alg}.

\begin{table*}[t]
\centering
\caption{The average test accuracy ($\%$) on CIFAR100N-C and CIFAR80N-O over the last 10 epochs. ``Sym'' and ``Asym'' denote the symmetric and asymmetric label noise, respectively. Experimental results of existing methods are mainly copied from \cite{sun2021josrc,sun2022pnp}. The symbol $^{\dagger}$ signifies that we re-implement the respective method using its open-sourced code and default hyper-parameter setting. \majrev{The best and the second-best results are bolded and underlined, respectively.}}
\begin{tabular}{@{}rcccccccc@{}}
\toprule
\multicolumn{1}{c}{\multirow{2}{*}{Methods}} & \multicolumn{4}{c}{CIFAR100N-C} & \multicolumn{4}{c}{CIFAR80N-O} \\ \cmidrule(l){2-9} 
\multicolumn{1}{c}{} & Sym-20\% & Sym-50\% & Sym-80\% & Asym-40\% & Sym-20\% & Sym-50\% & Sym-80\% & Asym-40\% \\ \midrule
Standard & 35.14 $\pm$ 0.44 & 16.97 $\pm$ 0.40 & 4.41 $\pm$ 0.14 & 27.29 $\pm$ 0.25 & 29.37 $\pm$ 0.09 & 13.87 $\pm$ 0.08 & 4.20 $\pm$ 0.07 & 22.25 $\pm$ 0.08 \\
Decoupling \cite{decoupling} & 33.10 $\pm$ 0.12 & 15.25 $\pm$ 0.20 & 3.89 $\pm$ 0.16 & 26.11 $\pm$ 0.39 & 43.49 $\pm$ 0.39 & 28.22 $\pm$ 0.19 & 10.01 $\pm$ 0.29 & 33.74 $\pm$ 0.26 \\
Co-teaching \cite{coteaching} & 43.73 $\pm$ 0.16 & 34.96 $\pm$ 0.50 & 15.15 $\pm$ 0.46 & 28.35 $\pm$ 0.25 & 60.38 $\pm$ 0.22 & 52.42 $\pm$ 0.51 & 16.59 $\pm$ 0.27 & 42.42 $\pm$ 0.30 \\
Co-teaching+ \cite{coteachingplus} & 49.27 $\pm$ 0.03 & 40.04 $\pm$ 0.70 & 13.44 $\pm$ 0.37 & 33.62 $\pm$ 0.39 & 53.97 $\pm$ 0.26 & 46.75 $\pm$ 0.14 & 12.29 $\pm$ 0.09 & 43.01 $\pm$ 0.59 \\
JoCoR \cite{wei2020combating} & 53.01 $\pm$ 0.04 & 43.49 $\pm$ 0.46 & 15.49 $\pm$ 0.98 & 32.70 $\pm$ 0.35 & 59.99 $\pm$ 0.13 & 50.61 $\pm$ 0.12 & 12.85 $\pm$ 0.05 & 39.37 $\pm$ 0.16 \\
\rev{TopoFilter $^{\dagger}$} \cite{neurips2020_topofilter} & 52.53 $\pm$ 0.18 & 39.59 $\pm$ 0.13 & 21.55 $\pm$ 0.67 & 50.11 $\pm$ 0.26 & 54.62 $\pm$ 0.28 & 41.21 $\pm$ 0.24 & 22.17 $\pm$ 0.89 & 51.35 $\pm$ 0.22 \\
\rev{FINE $^{\dagger}$} \cite{neurips2021_FINE} & 60.35 $\pm$ 0.07 & 50.97 $\pm$ 0.06 & 25.82 $\pm$ 0.07 & 49.29 $\pm$ 0.12 & 61.43 $\pm$ 0.12 & 52.23 $\pm$ 0.07 & 26.14 $\pm$ 0.06 & 48.68 $\pm$ 0.15 \\
Jo-SRC \cite{sun2021josrc} & 58.15 $\pm$ 0.14 & 51.26 $\pm$ 0.11 & 23.80 $\pm$ 0.05 & 38.52 $\pm$ 0.20 & 65.83 $\pm$ 0.13 & 58.51 $\pm$ 0.08 & 29.76 $\pm$ 0.09 & 53.03 $\pm$ 0.25 \\
Co-LDL \cite{sun2021coldl} & 59.73 $\pm$ 0.11 & 49.52 $\pm$ 0.24 & 25.12 $\pm$ 0.21 & 52.28 $\pm$ 0.09 & 58.81 $\pm$ 0.19 & 48.64 $\pm$ 0.35 & 24.22 $\pm$ 0.17 & 50.69 $\pm$ 0.20 \\
PNP-hard \cite{sun2022pnp} & \underline{64.25 $\pm$ 0.12} & 54.37 $\pm$ 0.13 & 30.26 $\pm$ 0.15 & 56.01 $\pm$ 0.31 & 65.87 $\pm$ 0.23 & 54.41 $\pm$ 0.30 & 30.79 $\pm$ 0.16 & 56.17 $\pm$ 0.42 \\
PNP-soft \cite{sun2022pnp} & 63.27 $\pm$ 0.14 & \underline{57.10 $\pm$ 0.16} & 31.32 $\pm$ 0.19 & 60.25 $\pm$ 0.21 & \underline{67.00 $\pm$ 0.18} & \underline{59.08 $\pm$ 0.24} & 34.36 $\pm$ 0.18 & \underline{61.23 $\pm$ 0.17} \\ 
UNICON $^{\dagger}$ \cite{karim_unicon_2022} & 55.10 $\pm$ 0.30 & 55.11 $\pm$ 0.23 & 31.49 $\pm$ 0.18 & 49.90 $\pm$ 0.30 & 54.50 $\pm$ 0.50 & 50.73 $\pm$ 0.17 & 36.75 $\pm$ 0.14 & 51.50 $\pm$ 0.50 \\
NCE $^{\dagger}$ \cite{avidan_neighborhood_2022} & 54.58 $\pm$ 0.51 & 53.17 $\pm$ 0.60 & \underline{35.23 $\pm$ 0.27} & 49.90 $\pm$ 0.54 & 58.53 $\pm$ 0.11 & 57.08 $\pm$ 0.23 & \underline{39.34 $\pm$ 0.15} & 56.40 $\pm$ 0.49 \\
SOP $^{\dagger}$ \cite{liu2022robust} & 58.63 $\pm$ 0.01 & 53.75 $\pm$ 0.11 & 34.23 $\pm$ 0.63 & 49.87 $\pm$ 0.44 & 60.17 $\pm$ 0.78 & 54.77 $\pm$ 0.21 & 34.05 $\pm$ 0.59 & 53.34 $\pm$ 0.43 \\
\majrev{NPN} $^{\dagger}$ \cite{sheng2024adaptive} & 62.76 $\pm$ 0.13 & 56.27 $\pm$ 0.15 & 31.69 $\pm$ 0.13 & 57.11 $\pm$ 0.26 & 63.78 $\pm$ 0.29 & 56.91 $\pm$ 0.20 & 25.25 $\pm$ 0.25 & 58.50 $\pm$ 0.34 \\ 
\majrev{CA2C}$^{\dagger}$ \cite{sheng2025ca2c}     & 63.56 $\pm$ 0.09 & 54.76 $\pm$ 0.06 & 33.56 $\pm$ 0.07 & \underline{61.16 $\pm$ 0.06} & 64.59 $\pm$ 0.10 & 56.86 $\pm$ 0.12 & 31.27 $\pm$ 0.14 & 60.19 $\pm$ 0.10 \\
\midrule
Jo-SNC & \textbf{65.49 $\pm$ 0.15} & \textbf{59.88 $\pm$ 0.17} & \textbf{43.45 $\pm$ 0.16} & \textbf{61.59 $\pm$ 0.10} & \textbf{67.81 $\pm$ 0.15} & \textbf{61.78 $\pm$ 0.15} & \textbf{41.10 $\pm$ 0.12} & \textbf{62.57 $\pm$ 0.19} \\ 
\bottomrule
\end{tabular}
\label{tab:cifar_benchmark}
\end{table*}

\begin{figure*}[!t]
\centering
\includegraphics[width=1\linewidth]{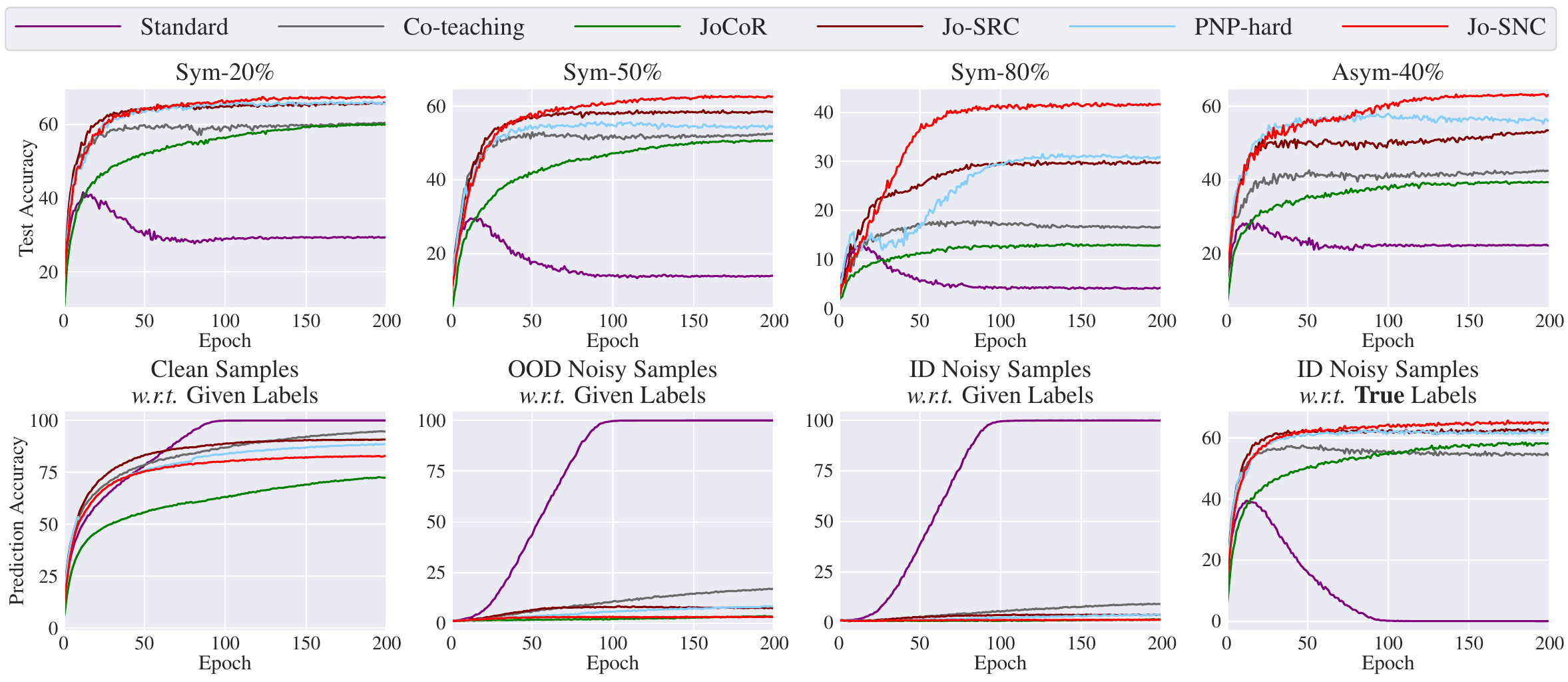}
\caption{
\textbf{(Top)} Comparison of test accuracy (\%) \vs epochs with state-of-the-art sample-selection-based approaches on CIFAR80N-O datasets under various noise conditions (\ie, Sym-20\%, Sym-50\%, Sym-80\%, and Asym-40\%). 
\textbf{(Bottom)} Comparison of prediction accuracy (\%) of different types of training samples (\ie, clean ones, ID noisy ones, and OOD noisy ones) on the CIFAR80N-O (Sym-20\%) dataset during the training process.
}
\label{fig:test_acc_and_split_acc}
\end{figure*}

\section{Experiments}
This section first introduces the experimental settings, including benchmark datasets, evaluation metrics, implementation details, and state-of-the-art (SOTA) baselines.
Then we compare Jo-SNC with SOTA methods on synthetic noisy datasets and real-world datasets for performance evaluation.
Lastly, we perform an in-depth analysis of Jo-SNC through comprehensive ablation studies.

\subsection{Experiment setup}
\subsubsection{Datasets}
We evaluate our approach on two synthetic noisy datasets (\ie, CIFAR100N-C, CIFAR80N-O) and six real-world noisy datasets (\ie, Web-Aircraft, Web-Bird, Web-Car \cite{sun2021webly}, Animal-10N \cite{song2019selfie}, WebVision \cite{li2017webvision}, and Food101N \cite{lee2018cleannet}).
Details of these benchmark datasets are illustrated as follows.

\textbf{CIFAR100N-C and CIFAR80N-O} are two synthetic datasets created from CIFAR100 \cite{krizhevsky2009}.
Specifically, we follow JoCoR \cite{wei2020combating} to create the closed-set synthetic dataset CIFAR100N-C with a noise ratio $\mathfrak{n}_{c} \in (0, 1)$. 
The noise type $\mathfrak{T}$ could be either ``Symmetric'' or ``Asymmetric''. 
Symmetric noise corrupts sample labels by randomly assigning them to all possible classes, while asymmetric noise mimics real-world label noise by corrupting sample labels only to their similar categories.
To create the open-set synthetic dataset CIFAR80N-O, we first regard the last 20 categories in CIFAR100 as out-of-distribution ones. 
Then we create in-distribution noisy samples by randomly corrupting $\mathfrak{n}_{c}$ percentage of remaining sample labels in a $\mathfrak{T}$ fashion. 
This finally leads to an overall noise ratio $\mathfrak{n}_{all} = 0.2+0.8\mathfrak{n}_{c}$.

\textbf{Web-Aircraft, Web-Bird, and Web-Car} are subsets of the web-image-based fine-grained image dataset \textit{WebFG-496} \cite{sun2021webly}.
They are three fine-grained datasets containing noisy training images crawled from web image search engines. 
Web-Aircraft is composed of 100 aircraft categories and contains 13503 noisy training images and 3333 test images. 
Web-Bird contains 18388 noisy training images and 5794 clean test images belonging to 200 bird species. 
Web-Car consists of 196 different car types that contain 21448 noisy training samples and 8041 clean test samples.

\begin{table*}[t]
\renewcommand\tabcolsep{12pt}
\centering
\caption{Comparison with state-of-the-art approaches in test accuracy ($\%$) on Web-Aircraft, Web-Bird, and Web-Car. Results of previous approaches are mostly from \cite{sun2022pnp}. The symbol $^{\dagger}$ means that we re-implement the method using its open-sourced code and default hyper-parameter setting. \majrev{The best and the second-best results are bolded and underlined, respectively.}}
\begin{tabular}{rcccccc}
\toprule
\multicolumn{1}{c}{\multirow{2}{*}{Methods}} & \multirow{2}{*}{Publications} & \multirow{2}{*}{Backbone} & \multicolumn{4}{c}{Performances ($\%$)} \\
\cmidrule{4-7}  &    &     & Web-Aircraft  &  Web-Bird  & Web-Car	&	Average \\
\midrule
Standard									&   -         & ResNet50 & 	60.80	&	64.40	&	60.60	&	61.93	\\
Decoupling \cite{decoupling}          		&NeurIPS 2017 & ResNet50 & 	75.91	&   71.61  	&   79.41	&	75.64	\\
Co-teaching \cite{coteaching}         		&NeurIPS 2018 & ResNet50 &	79.54	&   76.68	&   84.95	&	80.39	\\
Co-teaching+ \cite{coteachingplus}			&ICML 2019    & ResNet50 &  74.80	&   70.12   &   76.77	&	73.90	\\
SELFIE \cite{song2019selfie}				&ICML 2019	  & ResNet50 &  79.27	&	77.20	&	82.90	&	79.79	\\
PENCIL \cite{yi2019probabilistic}			&CVPR 2019    & ResNet50 &  78.82   &   75.09   &   81.68	&	78.53	\\
JoCoR \cite{wei2020combating}				&CVPR 2020    & ResNet50 &  80.11   &   79.19   &   85.10	&	81.47	\\
AFM \cite{peng2020suppressing}				&ECCV 2020    & ResNet50 &  81.04   &   76.35   &   83.48	&	80.29	\\
CRSSC \cite{sun2020crssc}					&ACM MM 2020  & ResNet50 &  82.51	&	81.31	&	87.68	&	83.83	\\
Self-adaptive \cite{huang2020self}			&NeurIPS 2020 & ResNet50 &  77.92   &   78.49   &   78.19	&	78.20	\\
DivideMix \cite{li2020dividemix}			&ICLR 2020	  & ResNet50 &  82.48	&	74.40	&	84.27	&	80.38	\\
\rev{TopoFilter $^{\dagger}$ }~\cite{neurips2020_topofilter} &NeurIPS 2020&ResNet50&  78.42	&   78.56	&   83.62   &   80.20   \\
\rev{FINE $^{\dagger}$ }~\cite{neurips2021_FINE}          &NeurIPS 2021 & ResNet50 &  80.83	&   77.75	&   85.39   &   81.32   \\
PLC \cite{zhang2021learning}				&ICLR 2021	  & ResNet50 &	79.24	&	76.22	&	81.87	&	79.11	\\
Peer-learning \cite{sun2021webly}			&ICCV 2021	  & ResNet50 &	78.64	&	75.37	&	82.48	&	78.83	\\
Jo-SRC \cite{sun2021josrc}					&CVPR 2021	  & ResNet50 &  82.73	&	81.22	&	88.13	&	84.03	\\
PNP-hard \cite{sun2022pnp}					&CVPR 2022    & ResNet50 &  85.03   &   81.20   &   89.93	&	85.39	\\
PNP-soft \cite{sun2022pnp}					&CVPR 2022    & ResNet50 & 	85.54   &   81.93   &   \underline{90.11}	&	85.86	\\
UNICON $^{\dagger}$ \cite{karim_unicon_2022} 		& CVPR 2022 & ResNet50 & 85.18 & 81.20 & 88.15 &		84.84	\\
NCE $^{\dagger}$ \cite{avidan_neighborhood_2022} 	& ECCV 2022 & ResNet50 & 84.94 & 80.22 & 86.38 &		83.85	\\
SOP $^{\dagger}$ \cite{liu2022robust} 				& ICML 2022 & ResNet50 & 84.06 & 79.40 & 85.71 &		83.06	\\
SPRL $^{\dagger}$ \cite{shi2023self}                & PR 2023   & ResNet50 & 84.40 & 80.27 & 87.61 &        84.09   \\
AGCE $^{\dagger}$ \cite{zhou2023asymmetric}         & TPAMI 2023& ResNet50 & 84.22 & 75.60 & 85.16 &        81.66   \\
\majrev{NPN} \cite{sheng2024adaptive}               & AAAI 2024 & ResNet50 & 83.65 & 79.36 & 85.46 &        82.82   \\
\majrev{CA2C} \cite{sheng2025ca2c}                  & ICCV 2025 & ResNet50 & \underline{87.70} & \underline{82.48} & 89.11 &        \underline{86.43}   \\
\midrule
Jo-SNC & - & ResNet50 & \textbf{87.97} & \textbf{82.76} & \textbf{90.56} & \textbf{87.10}	\\
\bottomrule	
\end{tabular}
\label{tab:webfg_benchmark}
\end{table*}

\textbf{Animal-10N \cite{song2019selfie}} is also a real-world noisy dataset. It contains 55k images belonging to 5 pairs of confusing animals. 
These images are crawled from online image search engines (\eg, Bing, Google) and then classified by 15 participants.
The overall noise ratio is about 8\%. 
Following \cite{song2019selfie}, 50k images are employed as training data while the other 5k instances are used for testing.

\textbf{WebVision \cite{li2017webvision}} is a large-scale real-world dataset associated with label noise.
This dataset contains 2.4 million images that are crawled from Flickr and Google using the 1000 categories in the ImageNet ILSVRC12 \cite{ILSVRC15}.
For ease of comparison with previous works, we employ the mini-WebVision dataset which contains the first 50 classes from the Google image subset of WebVision \cite{shu2023cmw}. The number of training samples is around 66k.
We adopt both the mini-WebVision validation set and the ImageNet ILSVRC12 validation set for evaluating our proposed method.

\textbf{Food101N \cite{lee2018cleannet}} is another large-scale real-world food dataset with noisy labels. 
The data is collected from Google, Bing, Yelp, and TripAdvisor \cite{lee2018cleannet} based on the category taxonomy of Food101 \cite{bossard14}.
Food101N contains around 310k training samples belonging to 101 types of food.
The accuracy of annotation is roughly 80\%.
Following the common experimental setting, we employ the 25k test images of Food101 as the test set for Food101N.

\subsubsection{Evaluation metrics}
For evaluating the model classification performance, we take test accuracy as the evaluation metric.
Reported results are averaged performance of five repeated experiments under identical settings.

\subsubsection{Implementation details}
\label{sec:implementation_details}
Experiments are implemented using the Pytorch framework. Detailed information for each dataset is as follows.

\textit{CIFAR100N-C and CIFAR80N-O}:
Following \cite{sun2021josrc}, we adopt a 7-layer DNN for CIFAR100N-C and CIFAR80N-O. 
During training, we use Adam optimizer with a momentum of 0.9 for network optimization. 
The initial learning rate is 0.001, and the batch size is 128. 
We train the network for 200 epochs and the first 10 epochs are warmup. 
We start to decay the learning rate after the warmup stage in the cosine annealing manner. 
The LSR parameter $\epsilon$ is empirically set to 0.6.
The default length of the queue $\mathscr{Q}$ is set to 32000 and the default embedding dimension is set to 256.
The default number of $K$ nearest neighbors is set to $K=10$ and the default value of $\kappa$ is set to $\kappa=5$.
Empirically, $\omega_{\tau}$ is set to 0.75 in warmup epochs.
In the subsequent robust training epochs, $\omega_{\tau}$ is set to 0.925 for Sym-80\%, 0,995 for Sym-20\%, and 0.975 otherwise.
Default hyper-parameters $\alpha$, $\beta$, and $\gamma$ are set to 0.3, 0.1, and 0.0001, respectively.

\textit{Web-Aircraft, Web-Bird, and Web-Car}:
Following \cite{sun2022pnp}, we employ ResNet50 \cite{he2016deep} pre-trained on ImageNet as our backbone for Web-Aircraft, Web-Bird, and Web-Car.
We adopt the SGD with a momentum of 0.9 as the optimizer. 
The initial learning rate and the batch size are set to 0.01 and 64, respectively.
The warmup stage lasts for 5 epochs and we train our network for 120 epochs in total.
The learning rate starts to decay after 10 epochs in a cosine annealing manner.
The LSR parameter $\epsilon$ is empirically set to 0.3.
The default length of the queue $\mathscr{Q}$ is set to 32000 and the default embedding dimension is set to 512. 
The default number of $K$ nearest neighbors is set to $K=10$. 
$\kappa$ is set to $\kappa=5$ for Web-Aircraft and $\kappa=10$ for the other two datasets.
Empirically, $\omega_{\tau}$ is set to 0.75 in warmup epochs and 0.98 in the subsequent robust training stage.
Default hyper-parameters $\alpha$, $\beta$, and $\gamma$ are set to 0.3, 0.8, and 0.001, respectively.

\textit{Animal-10N}:
Following \cite{shu2023cmw}, we employ VGG19 with BatchNorm \cite{simonyan2014very} as the backbone. 
The initial learning rate and the batch size are set to 0.01 and 128, respectively. 
The network is trained for 100 epochs with a 5-epoch warmup.
The LSR parameter $\epsilon$ is set to 0.8.
Default hyper-parameters $\alpha$, $\beta$, and $\gamma$ are set to 0.2, 0.1, and 0.005, respectively.
Other than the above parameters, the remaining setting is similar to that used for Web-Aircraft.

\textit{mini-WebVision}:
Following \cite{shu2023cmw}, we employ ResNet50 \cite{he2016deep} and Inception-ResNet-v2 \cite{szegedy2017inception} as backbones.
The network is trained for 100 epochs, in which the warmup lasts for 5 epochs.
The learning rate drops by a factor of 0.1 at the 30-th, 60-th and 80-th epoch.
$\epsilon$ is set to 0.5. $\alpha$, $\beta$, and $\gamma$ are set to 0.2, 0.1, and 0.005, respectively.
Other settings are similar to Web-Aircraft.

\textit{Food101N}:
Following \cite{sun2021josrc}, we use ResNet50 \cite{he2016deep} pre-trained on ImageNet as the backbone. 
We use SGD with a momentum of 0.9 to optimize our network. 
The initial learning rate and the batch size are set to 0.01 and 64, respectively.
We train our network for 60 epochs and the first 5 epochs are warmup.
The learning rate drops by a factor of 0.1 every 10 epochs after 15 epochs of training.
The LSR parameter $\epsilon$ is empirically set to 0.65.
$\kappa$ is set to $\kappa=5$. $\omega_{\tau}$ is set to 0.75 in warmup epochs and 0.99 in the subsequent epochs.
Default hyper-parameters $\alpha$, $\beta$, and $\gamma$ are set to 0.8, 0.1, and 0.001, respectively.

\subsubsection{Baselines}
To evaluate our proposed approach, besides adopting Standard (\ie, training directly using noisy datasets) as a simple baseline, we compare Jo-SNC with state-of-the-art methods to demonstrate its effectiveness and superiority.

For \textit{CIFAR100N-C} and \textit{CIFAR80N-O}, we compare Jo-SNC with the following state-of-the-art methods: Decoupling \cite{decoupling}, Co-teaching \cite{coteaching}, Co-teaching+ \cite{coteachingplus}, JoCoR \cite{wei2020combating}, \rev{TopoFilter~\cite{neurips2020_topofilter}, FINE~\cite{neurips2021_FINE},} Jo-SRC \cite{sun2021josrc}, Co-LDL \cite{sun2021coldl}, PNP \cite{sun2022pnp}, UNICON \cite{karim_unicon_2022}, NCE \cite{avidan_neighborhood_2022}, SOP \cite{liu2022robust}, \majrev{NPN \cite{sheng2024adaptive}, and CA2C \cite{sheng2025ca2c}}.

For \textit{Web-Aircraft}, \textit{Web-Bird}, and \textit{Web-Car}, the comparison state-of-the-art methods include: Decoupling \cite{decoupling}, Co-teaching \cite{coteaching}, Co-teaching+ \cite{coteachingplus}, SELFIE \cite{song2019selfie}, PENCIL \cite{yi2019probabilistic}, JoCoR \cite{wei2020combating}, AFM \cite{peng2020suppressing}, CRSSC \cite{sun2020crssc}, Self-adaptive \cite{huang2020self}, DivideMix \cite{li2020dividemix}, \rev{TopoFilter~\cite{neurips2020_topofilter}, FINE~\cite{neurips2021_FINE},} PLC \cite{zhang2021learning}, Peer-learning \cite{sun2021webly}, Jo-SRC \cite{sun2021josrc}, PNP \cite{sun2022pnp}, UNICON \cite{karim_unicon_2022}, NCE \cite{avidan_neighborhood_2022}, SOP \cite{liu2022robust}, SPRL \cite{shi2023self}, AGCE \cite{zhou2023asymmetric}, \majrev{NPN \cite{sheng2024adaptive}, and CA2C \cite{sheng2025ca2c}}.

For \textit{Animal-10N}, ActiveBias \cite{chang2017active}, Co-teaching \cite{coteaching}, SELFIE \cite{song2019selfie}, PLC \cite{zhang2021learning}, MW-Net \cite{shu2019meta}, CMW-Net \cite{shu2023cmw}, \majrev{and AdaptCDRP} \cite{nips2024adaptcdr} are compared.

For \textit{mini-WebVision}, competing methods include: D2L \cite{ma2018dimensionality}, F-correction \cite{patrini2017making}, MentorNet \cite{mentornet}, Co-teaching \cite{coteaching}, Interative-CV \cite{chen2019understanding}, DivideMix \cite{li2020dividemix}, ELR \cite{liu2020early}, UNICON \cite{karim_unicon_2022}, NCE \cite{avidan_neighborhood_2022}, SOP \cite{liu2022robust}, MW-Net \cite{shu2019meta}, CMW-Net and its variants \cite{shu2023cmw}, \majrev{DivideMix-GM and C2D-GM \cite{garg2024ccgm}}.

For \textit{Food101N}, CleanNet \cite{lee2018cleannet}, MetaCleaner \cite{zhang2019metacleaner}, DeepSelf \cite{han2019deep}, AFM \cite{peng2020suppressing}, PLC \cite{zhang2021learning}, Jo-SRC \cite{sun2021josrc}, PNP \cite{sun2022pnp}, and \majrev{MultiProto~\cite{zhang2025learning}}  are compared with our approach.

\subsection{Experiments on synthetic noisy datasets}

\subsubsection{Results on CIFAR100N-C}
Although this work primarily aims to cope with open-set noisy labels, we find that our proposed method also demonstrates effectiveness in closed-set noisy scenarios. 
To substantiate the efficacy of our method in closed-set noisy datasets, we provide experimental results on CIFAR100N-C under various conditions (\eg, different noise structures and noise rates) in Table~\ref{tab:cifar_benchmark}.
By comparing the proposed Jo-SNC with existing state-of-the-art approaches, we can find that our method consistently obtains leading performance in various noisy situations.
Compared with \cite{sun2021josrc}, Jo-SNC achieves significant performance improvement under all noisy conditions (\ie, 7.34\%, 8.62\%, 19.65\%, and 23.07\% performance boost on Sym-20\%, Sym-50\%, Sym-80\%, and Asym-40\%, respectively). 
Especially, Jo-SNC is empowered with remarkably stronger capability in tackling more challenging noisy scenarios (\ie, Sym-80\% and Asym-40\%) by incorporating self-consistency and neighbor-consistency.
Contrasting our approach with the latest state-of-the-art approaches (\ie, \cite{sun2022pnp,karim_unicon_2022,avidan_neighborhood_2022,liu2022robust,sheng2024adaptive,sheng2025ca2c}), our method also outperforms them consistently, demonstrating the robustness and superiority of our proposed approach.
It is worth noting that while the performance of all methods experiences a significant drop in the most inferior scenario (\ie, Sym-80\%), our method still secures the highest test accuracy.

\begin{table}[t]
\renewcommand\tabcolsep{15pt}
\centering
\caption{Comparison with state-of-the-art methods in test accuracy ($\%$) on Animal-10N. Results of existing methods are copied from \cite{shu2023cmw,nips2024adaptcdr}. \majrev{The best and the second-best results are bolded and underlined, respectively.}}
\begin{tabular}{@{}rcc@{}}
\toprule
\multicolumn{1}{c}{Methods} & Backbone & Test Accuracy \\ 
\midrule
Standard 							& VGG19\_BN & 79.40 $\pm$ 0.14 \\
ActiveBias \cite{chang2017active} 	& VGG19\_BN & 80.50 $\pm$ 0.26 \\
Co-teaching \cite{coteaching} 		& VGG19\_BN & 80.20 $\pm$ 0.13 \\
SELFIE \cite{song2019selfie} 		& VGG19\_BN & 81.80 $\pm$ 0.09 \\
PLC \cite{zhang2021learning} 		& VGG19\_BN & 83.40 $\pm$ 0.43 \\
MW-Net \cite{shu2019meta} 			& VGG19\_BN & 80.70 $\pm$ 0.52 \\
CMW-Net \cite{shu2023cmw} 			& VGG19\_BN & 80.90 $\pm$ 0.48 \\
CMW-Net-SL \cite{shu2023cmw} 		& VGG19\_BN & \underline{84.70 $\pm$ 0.28} \\
\majrev{AdaptCDRP} \cite{nips2024adaptcdr} & VGG19\_BN & 83.08 $\pm$ 0.39 \\
\midrule
Jo-SNC		                        & VGG19\_BN & \textbf{86.17 $\pm$ 0.11} \\
\bottomrule
\end{tabular}
\label{tab:animal10n_benchmark}
\end{table}

\subsubsection{Results on CIFAR80N-O}
As mentioned above, our proposed method is designed for learning with open-set noisy labels. 
Accordingly, we create CIFAR80N-O, a synthetic noisy dataset, to simulate the real-world scenarios (\ie, open-set noisy scenarios).
Table~\ref{tab:cifar_benchmark} presents the experimental comparison between our proposed method and existing state-of-the-art approaches on CIFAR80N-O under various noise conditions.
From this table, we can observe that our proposed Jo-SNC method outperforms existing state-of-the-art methods significantly and consistently.
When dealing with the easiest case (\ie, Sym-20\%), although existing methods demonstrate generally adequate performance, our method achieves the best test accuracy.
As the noise scenarios become harder (\eg, Sym-50\%), while the performance of all approaches starts to drop inevitably, our method still obtains the leading performance.
The performance advancement shown in this table demonstrates that our method can effectively handle not only symmetric label noise but also asymmetric one (\ie, Asym-40\%).
Lastly, in the most challenging scenario (\ie, Sym-80\%) where a large proportion of samples are associated with corrupted labels, all methods exhibit a significant performance drop. 
While most state-of-the-art methods can only achieve $< 35\%$ test accuracy, our proposed Jo-SNC surpasses existing methods by obtaining 41.10\% accuracy on test data. 
The performance improvement is 1.76\% compared to the second-best performer \cite{avidan_neighborhood_2022}. 
This firmly illustrates the superiority of our method in addressing noisy labels, including extremely noisy ones.
Figure~\ref{fig:test_acc_and_split_acc} (top) presents the test accuracy \vs epochs for models trained on CIFAR80N-O using different state-of-the-art sample-selection-based methods.
Notably, we can observe that our proposed Jo-SNC consistently outperforms these counterparts. 
Moreover, the superiority in the robustness of our method is also evidenced by the monotonically increasing trends of these curves.

\begin{table}[t]
\centering
\caption{Comparison with state-of-the-art methods in test accuracy ($\%$) on mini-WebVision. Results of existing methods are from their respective papers. Inception-ResNet-v2 (denoted as ``IRNv2'') and ResNet50 (denoted as ``RN50'') are both employed for model evaluation. \majrev{The best and the second-best results are bolded and underlined, respectively.}}
\begin{tabular}{@{}rccccc@{}}
\toprule
\multicolumn{1}{c}{\multirow{2}{*}{Methods}} & \multirow{2}{*}{Backbone} & \multicolumn{2}{c}{WebVision} & \multicolumn{2}{c}{ILSVRC12} \\ \cmidrule(l){3-6} 
 &  & top-1 & top-5 & top-1 & top-5 \\ \midrule
D2L \cite{ma2018dimensionality} 			& IRNv2 & 62.68 & 84.00 & 57.80 & 81.36 \\
F-correction \cite{patrini2017making} 		& IRNv2 & 61.12 & 82.68 & 57.36 & 82.36 \\
MentorNet \cite{mentornet} 					& IRNv2 & 63.00 & 81.40 & 57.80 & 79.92 \\
Co-teaching \cite{coteaching} 				& IRNv2 & 63.58 & 85.20 & 61.48 & 84.70 \\
Interative-CV \cite{chen2019understanding} 	& IRNv2 & 65.24 & 85.34 & 61.60 & 84.98 \\
DivideMix \cite{li2020dividemix} 			& IRNv2 & 77.32 & 91.64 & 75.20 & 90.84 \\
ELR \cite{liu2020early} 					& IRNv2 & 76.26 & 91.26 & 68.71 & 87.84 \\
ELR+ \cite{liu2020early} 					& IRNv2 & 77.78 & 91.68 & 70.29 & 89.76 \\
UNICON \cite{karim_unicon_2022} 			& IRNv2 & 77.60 & 93.44 & 75.29 & 93.72 \\
NCE \cite{avidan_neighborhood_2022} 		& IRNv2 & \underline{79.50} & \underline{93.80} & \underline{76.30} & \textbf{94.10} \\
SOP \cite{liu2022robust} 					& IRNv2 & 76.60 &   -   & 69.10 &   -   \\
\midrule
MW-Net \cite{shu2019meta} 					& RN50 	& 69.34 & 87.44 & 65.80 & 87.52 \\
DivideMix \cite{li2020dividemix} 			& RN50 	& 76.32 & 90.65 & 74.42 & 91.21 \\
DivideMix+C2D \cite{shu2023cmw} 			& RN50 	& 79.42 & 92.32 & 78.57 & 93.04 \\
CMW-Net \cite{shu2023cmw} 					& RN50 	& 70.56 & 88.76 & 66.44 & 87.68 \\
CMW-Net-SL \cite{shu2023cmw} 				& RN50 	& 78.08 & 92.96 & 75.72 & 92.52 \\
CMW-Net-SL+C2D \cite{shu2023cmw} 			& RN50 	& \underline{80.44} & \underline{93.36} & 77.36 & \underline{93.48} \\
\majrev{DivideMix-GM} \cite{garg2024ccgm}   & RN50  & 78.51 & 92.03 & 76.11 & 93.24 \\
\majrev{C2D-GM} \cite{garg2024ccgm}         & RN50  & 80.20 & 92.82 & \underline{79.16} & 93.12 \\
\midrule
Jo-SNC		& IRNv2 & \textbf{82.60} & \textbf{94.32} & \textbf{80.08} & \underline{93.84} \\
Jo-SNC		& RN50 & \textbf{82.32} & \textbf{93.68} & \textbf{80.16} & \textbf{93.88}  \\
\bottomrule
\end{tabular}
\label{tab:mimi_webvision_benchmark}
\end{table}

\begin{figure*}[t]
\centering
\includegraphics[width=\linewidth]{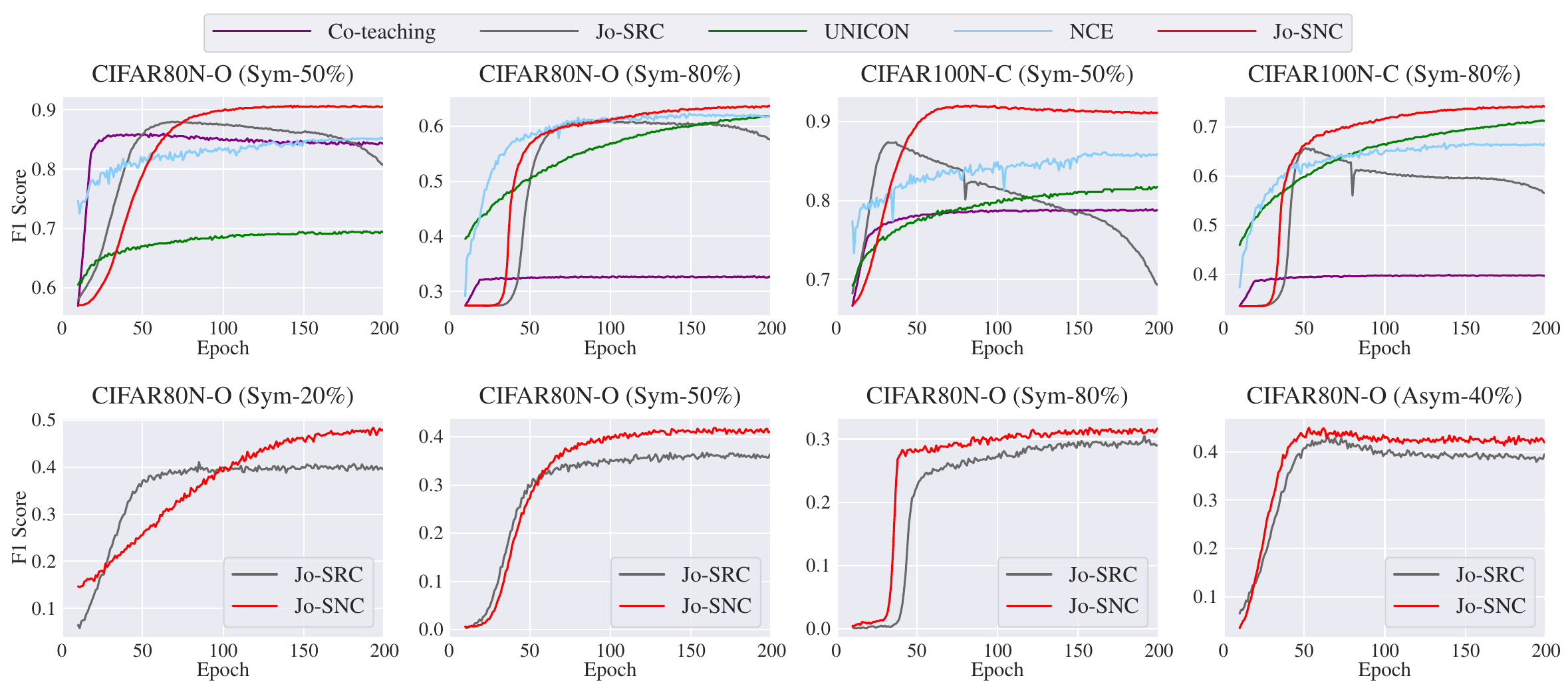}
\caption{
\textbf{(Top)} \rev{Comparison of (clean) sample selection performance between state-of-the-art methods and Jo-SNC on various noisy datasets.}
\textbf{(Bottom)} Comparison of (OOD noisy) sample selection performance between Jo-SRC \cite{sun2021josrc} and Jo-SNC on various noisy datasets.
The sample selection performance is evaluated using the F1 Score metric. 
These models are all trained for 200 epochs and the first 10 epochs are warmup. The evaluation of sample selection performance starts from the 11-th epoch.
}
\label{fig:selection_accuracy}
\end{figure*}


To further illustrate the effectiveness of our proposed method, we exhibit the prediction accuracy of different types of training samples (\ie, clean ones, ID noisy ones, and OOD noisy ones) in Figure~\ref{fig:test_acc_and_split_acc} (bottom). 
Additionally, we provide the prediction accuracy of ID noisy samples \wrt their ground-truth labels to help understand the training dynamics better.
From curves of ``Standard'' shown in Figure~\ref{fig:test_acc_and_split_acc} (bottom), we can see that while the prediction accuracy of clean data demonstrably increases during training, the prediction accuracy of noisy training samples (\wrt their given labels that are potentially corrupted) presents a similar trend. 
Oppositely, the prediction accuracy of ID noisy samples \wrt their ground-truth labels initially increases but then declines to nearly zero.  
This observation justifies the memorization effect \cite{zhang2016understanding} which argues that deep networks tend to fit clean and simple patterns before memorizing all samples (including noisy ones). 
However, this phenomenon is not desired in the training progress. Thus, it is critical to prevent networks from overfitting noisy labels when training with label-corrupted datasets.
As shown in these figures, existing methods all achieve increasing prediction accuracy on clean samples while obtaining low prediction accuracy on noisy ones (\wrt given labels). This indicates existing noise-robust methods are relatively effective in preventing overfitting.
Although Co-teaching \cite{coteaching} outperforms other methods in the prediction accuracy of clean samples, it suffers from an overfitting issue on noisy data.
Oppositely, while JoCoR \cite{wei2020combating} achieves low prediction accuracy on noisy samples, it yields an under-fitting issue on clean samples.
Jo-SRC \cite{sun2021josrc} and PNP-hard \cite{sun2022pnp} both exhibit better performance compared to our Jo-SNC in terms of prediction accuracy on clean samples. 
However, our Jo-SNC has higher robustness when learning from ID and OOD noisy samples. 
In particular, it achieves the lowest prediction accuracy on noisy samples (\wrt given labels) while obtaining the highest accuracy on ID noisy samples (\wrt true labels).
This evidently proves the effectiveness of our method, given that ground-truth labels of ID noisy samples are not provided as supervision during training.

\begin{table}[t]
\renewcommand\tabcolsep{15pt}
	\centering
    \caption{Comparison with state-of-the-art methods in test accuracy ($\%$) on Food101N using ResNet50 as the backbone. The results of prior methods are from their original papers. \majrev{The best and the second-best results are bolded and underlined, respectively.}}
	\begin{tabular}{rcc}
		\toprule
		\multicolumn{1}{c}{Methods}							&	Backbone	&	Test accuracy	\\
		\midrule
		Stardard											&	ResNet50	&	84.51			\\
		CleanNet $\omega_{hard}$ \cite{lee2018cleannet}		&	ResNet50	&	83.47			\\
		CleanNet $\omega_{soft}$ \cite{lee2018cleannet}		&	ResNet50	&	83.95			\\
		MetaCleaner \cite{zhang2019metacleaner}				&	ResNet50	&	85.05			\\
		DeepSelf	 \cite{han2019deep}						&	ResNet50	&	85.11			\\
		AFM \cite{peng2020suppressing}						&	ResNet50	&	87.23			\\
		PLC \cite{zhang2021learning}						&	ResNet50	&	85.28			\\
		Jo-SRC \cite{sun2021josrc}							&	ResNet50	&	86.66			\\
		PNP-hard \cite{sun2022pnp}							&	ResNet50	&	87.31			\\
		PNP-soft \cite{sun2022pnp} 							&	ResNet50	&	\underline{87.50}			\\
        \majrev{MultiProto}  \cite{zhang2025learning}       &   ResNet50    &   87.35           \\
		\midrule
		Jo-SNC		&	ResNet50	&	\textbf{88.02}	\\
		\bottomrule
	\end{tabular}
	\label{tab:food101n_benchmark}
\end{table}

\subsection{Experiments on real-world noisy datasets}

\subsubsection{Results on Web-Aircraft, Web-Bird, and Web-Car}
Beyond experimental comparisons on synthetic noisy datasets, we conduct model evaluations on real-world noisy datasets.
Web-Aircraft, Web-Bird, and Web-Car are three real-world fine-grained datasets whose training samples are crawled from web image search engines. 
Due to the nature of fine-grained categories, label noise in these three datasets is complicated. 
It is validated that at least 25\% of training annotations are associated with label corruptions, making it a practical and challenging task to train robust models.
Table~\ref{tab:webfg_benchmark} exhibits the comparison between our method and state-of-the-art methods on these three datasets.
We can observe that our proposed method consistently outperforms existing state-of-the-art methods by considerable margins. 
\majrev{In particular, our method achieves 0.27\%, 0.28\%, and 1.45\% performance gains compared to the second-best method (\ie, \cite{sheng2025ca2c}) on Web-Aircraft, Web-Bird, and Web-Car, respectively. Accordingly, the average performance improvement is 0.67\%.}
\rev{Notably, our approach significantly surpasses its counterparts that are also based on neighbors or grouping behaviors in the feature space (\ie, \cite{neurips2020_topofilter,neurips2021_FINE}), clearly demonstrating its superiority.}
These experimental results evidently verify the effectiveness of our method for coping with real-world datasets associated with such complicated label noise.

\begin{table*}
\centering
\renewcommand\tabcolsep{6pt}
\caption{
Ablation study about different variants of each component. 
Best results are marked with {\colorbox{mygray}{gray}}, denoting the default settings of Jo-SNC.
Experiments are all conducted on CIFAR80N-O with Sym-50\% and Asym-40\% label noise.
}
\renewcommand\thetable{6}
\subfloat[{Sample selection criterion} \label{tab:different_selection_criterion}]{
    \centering
    \begin{tabular}{ccc}
        \toprule
        \multirow{2}[3]{*}{\shortstack[c]{Selection \\ Criterion}} & \multicolumn{2}{c}{Test Accuracy} \\ \cmidrule(l){2-3} 
                    & Sym50 & Asym40 \\ 
        \midrule
        SelfOnly    & 60.92 & 62.40 \\
        NeiOnly     & 60.40 & 62.26 \\
        Dis         & 61.39 & 62.65 \\
        \rowcolor{mygray}
        SelfNeiDiv  & 62.70 & 63.06 \\ 
        \bottomrule
    \end{tabular}
}
\subfloat[{Sample selection threshold} \label{tab:different_selection_thresholding}]{
    \centering
    \begin{tabular}{ccc}
        \toprule
        \multirow{2}[3]{*}{\shortstack[c]{Threshold \\ Scheme}} & \multicolumn{2}{c}{Test Accuracy} \\ \cmidrule(l){2-3} 
                        & Sym50 & Asym40 \\ 
        \midrule
        Pre-defined     & 59.91 & 57.74 \\
        GMM             & 60.64 & 59.81 \\
        Mean            & 60.92 & 60.12 \\
        \rowcolor{mygray}
        Per-class       & 62.70 & 63.06 \\ 
        \bottomrule
    \end{tabular}
}
\subfloat[{ID noisy samples} \label{tab:different_learning_for_id}]{
    \centering
    \begin{tabular}{ccc}
        \toprule
        \multirow{2}[3]{*}{\shortstack[c]{Learning \\ Strategy}} & \multicolumn{2}{c}{Test Accuracy} \\ \cmidrule(l){2-3} 
                    & Sym50 & Asym40 \\ 
        \midrule
        Drop    & 59.55 & 60.19 \\
        SSL     & 61.84 & 62.72 \\
        \rowcolor{mygray}
        PLL     & 62.70 & 63.06 \\
        NL      & 61.19 & 62.12 \\ 
        \bottomrule
    \end{tabular}
}
\subfloat[{OOD noisy samples} \label{tab:different_learning_for_ood}]{
    \centering
    \begin{tabular}{ccc}
        \toprule
        \multirow{2}[3]{*}{\shortstack[c]{Learning \\ Strategy}} & \multicolumn{2}{c}{Test Accuracy} \\ \cmidrule(l){2-3} 
                    & Sym50 & Asym40 \\ 
        \midrule
        Drop    & 59.86 & 60.34 \\
        SSL     & 61.07 & 62.15 \\
        PLL     & 60.34 & 62.00 \\
        \rowcolor{mygray}
        NL      & 62.70 & 63.06 \\ 
        \bottomrule
    \end{tabular}
}
\label{tab:selection_threshold_idlearning_oodlearning}
\end{table*}

\begin{figure*}
\begin{minipage}{0.49\linewidth}
\centering
\includegraphics[width=\linewidth]{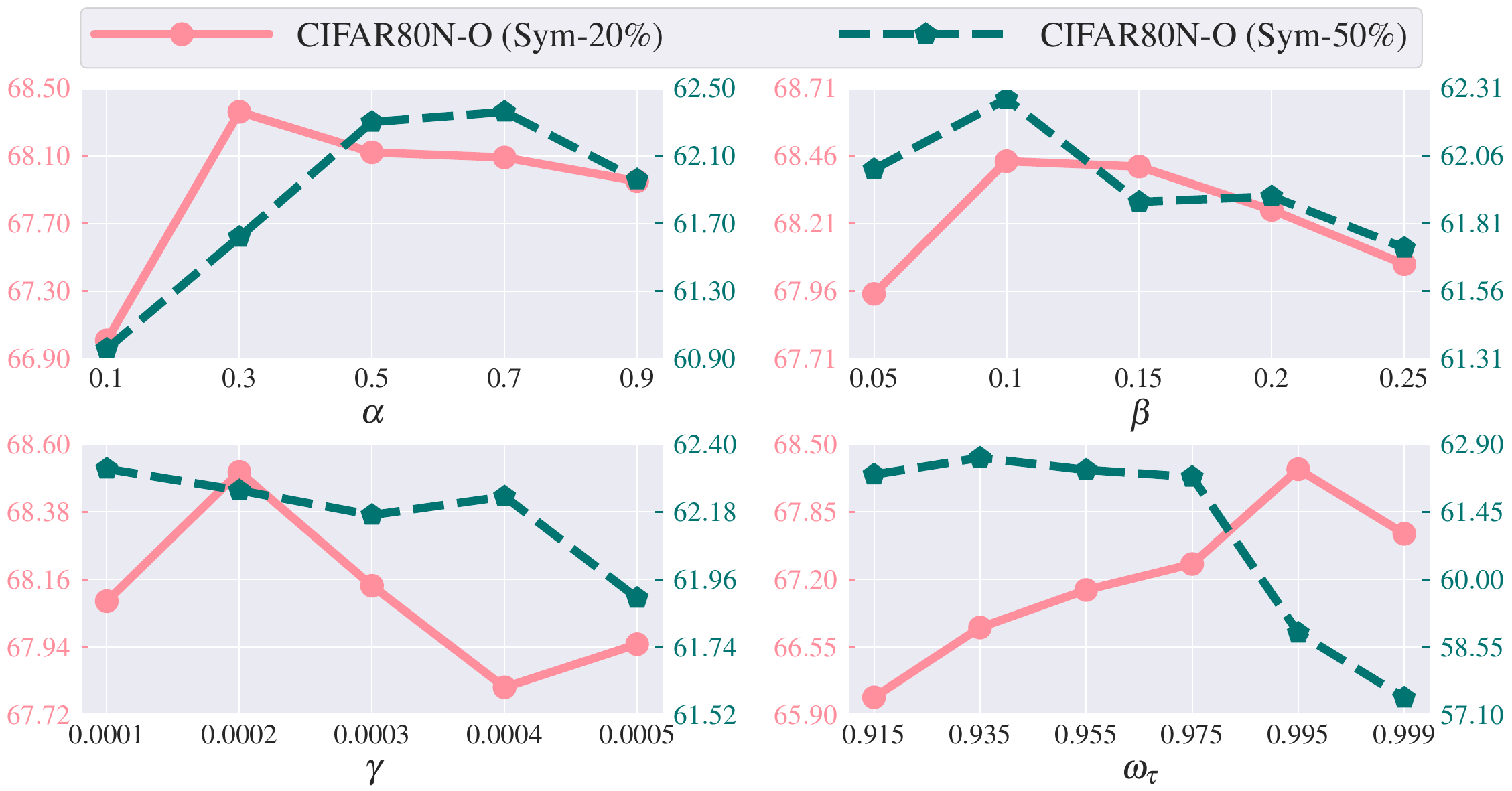}
\caption{Sensitivity of hyper-parameters $\alpha$, $\beta$, $\gamma$, and the threshold updating factor $\omega_{\tau}$. Experiments are conducted on CIFAR80N-O with Sym-20\% and Sym-50\% label noise.}
\label{fig:hyperparams}
\end{minipage}
\hspace{0.15cm}
\begin{minipage}{0.49\linewidth}
\centering
\includegraphics[width=\linewidth]{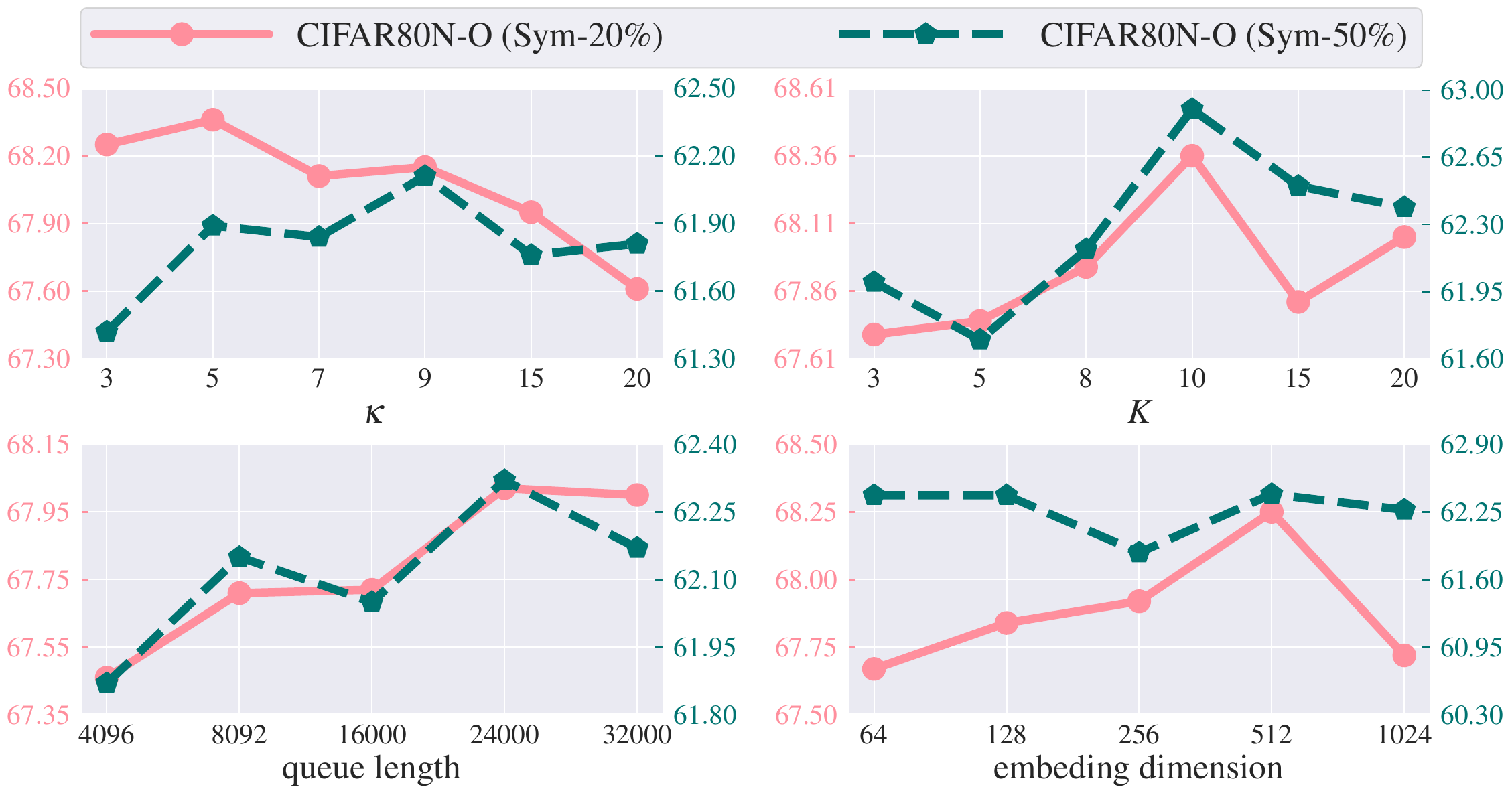}
\caption{Sensitivity of hyper-parameters $\kappa$, $K$, the queue length, and the embedding dimension. Experiments are conducted on CIFAR80N-O with Sym-20\% and Sym-50\% label noise.}
\label{fig:hyperparams-minor}
\end{minipage}
\end{figure*}

\subsubsection{Results on Animal-10N}
Table~\ref{tab:animal10n_benchmark} compares the test accuracy between our method and other competing methods on the Animal-10N dataset.
Compared with sample selection methods (\ie, ActiveBias \cite{chang2017active} and Co-teaching \cite{coteaching}), it is evident that our method attains better performance, validating the effectiveness of our sample selection strategy.
Our method also surpasses two competing label correction methods (\ie, SELFIE \cite{song2019selfie} and PLC \cite{zhang2021learning}).
Moreover, we compete with two sample re-weighting methods (\ie, MW-Net \cite{shu2019meta} and CMW-Net \cite{shu2023cmw}).
We can see that our method outperforms the latest CMW-Net \cite{shu2023cmw} method by 1.47\%.
This comparison validates the superiority of our method in coping with real-world noisy labels.


\subsubsection{Results on mini-WebVision}
Experimental results on mini-WebVision are revealed in Table~\ref{tab:mimi_webvision_benchmark}.
Both Inception-ResNet-v2 (denoted as ``IRNv2'') and ResNet50 (denoted as ``RN50'') are employed as backbones for evaluation.
Our method is accordingly compared with various state-of-the-art methods by using identical backbones.
It can be witnessed that when ResNet50 is adopted, our method consistently outperforms existing methods, especially in top-1 performance.
Jo-SNC obtains a 1.88\% / 1.59\% performance boost in top-1 accuracy on the validation set of WebVision / ImageNet ILSVRC12.
When we employ Inception-ResNet-v2 as the backbone, NCE \cite{avidan_neighborhood_2022} achieves 94.10\% in top-5 accuracy on the validation set of ImageNet ILSVRC12 (\ie, 0.22\% higher than ours).
However, our Jo-SNC still obtains the highest performance in the other cases (top-1 on both and top-5 on WebVision), demonstrating the superiority of our method.

\subsubsection{Results on Food101N}
Food101N is a large-scale real-world dataset with a large proportion of noisy labels. 
Table~\ref{tab:food101n_benchmark} illustrates the performance comparison between our method and state-of-the-art ones.
As shown in Table~\ref{tab:food101n_benchmark}, our Jo-SNC undoubtedly attains leading performance compared to these competing methods, surpassing the next-best one (\ie, PNP-soft \cite{sun2022pnp}) by 0.52\%.
This justifies that our proposed method can practically alleviate label noise when applied to large-scale noisy datasets in real-world scenarios.

\subsection{Ablation study}

\subsubsection{Analysis of sample selection}
Reliable sample selection is the key to our approach to achieving state-of-the-art performance.
To study and verify the superiority of our proposed sample selection strategy, we employ the F1 score to comprehensively evaluate the performance of sample selection. 
\rev{We provide the comparison between our method and multiple state-of-the-art methods (\ie, Co-teaching \cite{coteaching}, Jo-SRC \cite{sun2021josrc}, UNICON \cite{karim_unicon_2022}, and NCE \cite{avidan_neighborhood_2022}) in Figure~\ref{fig:selection_accuracy}. 
Notably, while all counterparts are compared when evaluating clean sample selection performance, our method is only compared with Jo-SRC for the OOD noisy sample selection performance because other methods do not differentiate between ID and OOD label noise.
This figure evidently demonstrates the effectiveness of our method in selecting clean and OOD noisy samples.
As present in Figure~\ref{fig:selection_accuracy}, our method consistently outperforms competing methods in both clean sample selection and OOD noisy sample selection in various noisy cases.
Regarding the clean sample selection, Jo-SRC \cite{sun2021josrc} shows a decrease in F1 scores in the later stage of training (especially in the most difficult case Sym-80\%). 
Contrarily, the F1 score of our method increases steadily throughout the training process.
Moreover, the results shown in Figure~\ref{fig:selection_accuracy} (top) effectively illustrate the superiority of our designed clean sample selection criterion (alongside our adaptive thresholding mechanism) compared to loss-based (Co-teaching), JS-distance-based (UNICON), and neighbor-based (NCE) sample selection methods.
}
In terms of OOD noise detection, our method reveals better detection performance than Jo-SRC \cite{sun2021josrc} although their F1 scores all keep increasing during training.

Table~\ref{tab:different_selection_criterion} presents the ablation study of employing different sample selection criteria.
``SelfOnly'' is the abbreviation of ``selecting clean samples using only condition (1) in Criterion~\ref{def:clean}. 
``NeiOnly'' denotes ``selecting clean samples using only condition (2) in Criterion~\ref{def:clean}. 
``Dis'' represents ``selecting OOD samples based on prediction disagreement''. 
``SelfNeiDiv'' means the proposed Criteria~\ref{def:clean} and \ref{def:id_ood} are employed for sample selection.
We can see that the integration of the two conditions in Criterion~\ref{def:clean} makes clean sample selection more reliable compared to using only one condition.
We can find that employing prediction divergence instead of disagreement for OOD noise detection achieves superior performance.
This comparison firmly demonstrates the effectiveness of our proposed sample selection strategy.


\subsubsection{Analysis of selection threshold}
We investigate different strategies for generating thresholds and present the comparison in Table~\ref{tab:different_selection_thresholding}.
``Pre-defined'' means that we adopt a linear schedule for thresholds with pre-defined initial and final values. 
``GMM'' represents that we resort to the Gaussian Mixture Model to generate unified thresholds. 
``Mean'' signifies that we obtain unified thresholds by computing mean values of $\mathcal{P}_{clean}$ and $\mathcal{P}_{ood}$. 
``Per-class'' abbreviates that we obtain class-specific thresholds by computing mean values of $\mathcal{P}_{clean}$ and $\mathcal{P}_{ood}$ for each class.
From this subfigure, we can observe that ``Per-class'' achieve the most superior performance, while the results of ``GMM'' and ``Mean'' are on par with each other.
We argue that the employment of class-specific thresholds offers advantages by alleviating imbalanced sample selection.
``Pre-defined'' obtains lower performance compared to its counterparts. 
Though this strategy could potentially attain comparable performance to other ones (or even higher than others) by meticulous parameter tuning, the pre-defined values are usually dataset-dependent. This inherent issue poses a challenge to the design of appropriate pre-defined values, ultimately leading to inferior generalization performance.

\subsubsection{Analysis of learning strategies for noisy samples}
Tables~\ref{tab:different_learning_for_id} and \ref{tab:different_learning_for_ood} show the effect of employing different learning strategies to leverage detected ID and OOD noisy samples.
``Drop'' means that detected noisy samples are neglected from training. 
``SSL'' signifies that we follow \cite{sun2021josrc} and employ a semi-supervised learning method to leverage selected noisy data for training. 
(Specifically, ID noisy samples are supervised by corrected labels estimated by a mean-teacher model, while OOD noisy samples are forced to fit an approximately uniform distribution.)
``PLL'' denotes that we resort to partial label learning for using selected noisy data. 
``NL'' represents that we employ negative learning to exploit detected noisy samples.
We can see that the model achieves the most inferior performance when detected ID / OOD noisy samples are discarded directly from training.
For ID noisy samples, it is evident that ``PLL'' attains the best performance.
We argue that this is because ground-truth labels of ID noisy samples are still within the known label space despite the difficulty in estimation. 
Therefore, adopting positive learning is more efficient than negative learning. 
Moreover, since the label re-assignment is hard to achieve accurately, using top-k predictions instead of the top-1 would encourage more robustness during training.
Contrarily, ``NL'' excels in learning from OOD noisy samples. 
We contend that the (out-of-distribution) ground-truth labels of OOD noisy samples pose a significant challenge in exploiting positive supervision effectively. 
Negative learning, by circumventing the need for positive labeling, offers more advantages in facilitating the learning of OOD noisy samples.

\begin{table}[t]
\centering
\caption{Impacts of different loss ingredients in test accuracy ($\%$) on CIFAR80N-O (Sym-50\%). Results at the best epochs are presented.}
\renewcommand\tabcolsep{5.5pt}
\begin{tabular}{ccccccc}
\toprule
Clean       &   ID        &   OOD      &   sCON      &   nCON     &   fCON	&	Test Accuracy   \\
\midrule
\checkmark  &             &            &            &            &    			& 56.11   \\
\checkmark  & \checkmark  &            &            &            &    			& 56.42   \\   
\checkmark  & \checkmark  & \checkmark &            &            &    			& 56.75   \\
\midrule   
\checkmark  & \checkmark  & \checkmark & \checkmark &            &    			& 59.66   \\   
\checkmark  & \checkmark  & \checkmark &            & \checkmark &				& 57.37   \\
\checkmark  & \checkmark  & \checkmark & 			& 			 & \checkmark	& 57.04   \\
\checkmark  & \checkmark  & \checkmark & \checkmark	& \checkmark	 &				& 60.79   \\
\checkmark  & \checkmark  & \checkmark & \checkmark	& 			 & \checkmark	& 61.29   \\   
\checkmark  & \checkmark  & \checkmark & 			& \checkmark	 & \checkmark	& 58.54   \\
\midrule   
\checkmark  & \checkmark  & \checkmark & \checkmark	& \checkmark	 & \checkmark	& 62.90   \\
\bottomrule        
\end{tabular}
\label{tab:different_modules}
\end{table}

\subsubsection{Analysis of impacts of different loss ingredients}
To exhibit insights on how effective each loss ingredient in our proposed Jo-SNC works, we conduct an ablation study on the impacts of different loss components of our method. 
Experiments are conducted on CIFAR80N-O (Sym-50\%) and results are shown in Table~\ref{tab:different_modules}.
``Clean'', ``ID'', and ``OOD'' denote the employment of Eq.~\eqref{eq:loss_cls} for different types of samples.
``sCON'', ``nCON'', and ``fCON'' are abbreviations for adopting Eqs.~\eqref{eq:loss_con_pred}, \eqref{eq:loss_con_neighbor}, and \eqref{eq:loss_con_feat}, respectively.
The first three rows in Table~\ref{tab:different_modules} illustrate the effect of using different types of samples (including clean, ID noisy, and OOD noisy ones). 
Given the low performance that ``Standard'' achieves, it is evident that our proposed clean sample selection plays the most crucial role in preventing the network from overfitting label noise.
Moreover, appropriate learning of ID noisy and OOD noisy samples can also boost the model performance.
The second group of experiments (\ie, the 4-th to the 9-th row) comprehensively explores the triplet consistency regularization.
The results reveal that while ``sCON'' makes the most significant contribution to the model when only one-level consistency is employed, combining these three consistency terms facilitates the most substantial improvement in model performance. 
This validates the necessity of our triplet consistency regularization.

\subsubsection{Analysis of hyper-parameter sensitivity}
For exploring the sensitivity of hyper-parameters in our method, we primarily investigate four major ones (\ie, $\alpha$, $\beta$, $\gamma$, and $\omega_{\tau}$).
Experiments are conducted on CIFAR80N-O with Sym-20\% and Sym-50\% label noise. Results are shown in Figure~\ref{fig:hyperparams}.
\rev{When one of these hyper-parameters is investigated, the other three are set using the default values specified in Section~\ref{sec:implementation_details}. (For example, when varying the value of $\alpha$ on CIFAR80N-O with Sym-20\% label noise, $\beta$ and $\gamma$ are set to 0.1 and 0.0001 while $\omega_{\tau}$ is set to 0.995.)}
$\alpha$, $\beta$, and $\gamma$ are weights to balance different loss terms in Eq.~\eqref{eq:loss_final}. 
$\alpha$ is studied in a value range of $\{0.1, 0.3, 0.5, 0.7, 0.9\}$. 
It can be seen that the best performance is obtained when $\alpha \in (0.3, 0.7)$. 
The value of $\beta$ and $\gamma$ varies in $\{0.05, 0.10, 0.15, 0.20, 0.25\}$ and $\{0.0001, 0.0002, 0.0003, 0.0004, 0.0005 \}$, respectively.
It is evident that our method is relatively robust against $\beta$ and $\gamma$.
Therefore, we set the default values of $\alpha$, $\beta$, and $\gamma$ as 0.3, 0.1, and 0.0001 for synthetic noisy datasets.
$\omega_{\tau}$ is the update factor for thresholds in Eq.~\eqref{eq:threshold}. 
A larger $\omega_{\tau}$ leads to a threshold that relies more on the past, while a smaller $\omega_{\tau}$ results in a threshold that depends more on the present.
The subfigure reveals that the model achieves the best performance on CIFAR80N-O (Sym-20\%) when $\omega_{\tau}$ is large (\ie, 0.995) while reaching the best performance on CIFAR80N-O (Sym-50\%) when $\omega_{\tau}$ is smaller (\ie, $\omega_{\tau} \le 0.975$).
Our hypothesis is that the generation of thresholds needs to rely more on the current values instead of the past ones when the label noise is heavier. 
If the label noise is severe, a value of $\omega_{\tau}$ that is too large would make the threshold generation slower than the model overfitting, resulting in sub-optimal performance.
On the contrary, when the noise rate is low, an overly small $\omega_{\tau}$ would increase the thresholds too fast, filtering more useful training samples and thus leading to under-learning.
Apart from these four major hyper-parameters, we additionally study four minor ones (\ie, the size of the partial label set $\kappa$, the number of nearest neighbors $K$, the queue length $\|\mathscr{Q}\|$, and the embedding dimension). Results are shown in Figure~\ref{fig:hyperparams-minor}. 
We can observe that Jo-SNC can perform marginally better if $\kappa \in [5,9]$ and $K \in [10, 20]$.
While a larger queue length or embedding dimension may lead to slightly higher performance, it would consume more memory space.
\majrev{We additionally conduct experiments on CIFAR80N-O (Sym-20\%) and CIFAR80N-O (Sym-50\%) by setting the queue length to 50k (\ie, the size of the entire training set). The results yield an average improvement of only 0.03\% and 0.09\%, respectively, compared to using a 32k queue. This provides evidence that our queue-based design (including both queue-based nearest-neighbor selection and feature consistency regularization) effectively minimizes computational overhead while delivering comparable performance.}
Overall, our method is considerably robust against these four parameters.



\section{Conclusion}
In this paper, we proposed a Jo-SNC algorithm aimed at mitigating performance deficiencies induced by noisy labels.
Jo-SNC started by distinguishing clean and noisy samples.
It selected clean samples globally according to JS divergence by simultaneously considering the samples themselves and their nearest neighbors.
ID and OOD noisy samples were distinguished based on prediction divergence between their different augmented views.
We introduced a data-driven and self-adaptive threshold generation strategy designed to adjust class-specific selection thresholds accordingly.
Subsequently, clean samples undertook conventional training, while ID and OOD noisy samples were learned through partial label learning and negative learning.
Finally, triplet consistency regularization, including self-prediction consistency, neighbor-prediction consistency, and feature consistency, was proposed to advance performance and robustness.
Through extensive experimentation and thorough ablation studies conducted on various synthetic and real-world noisy datasets, we substantiated the effectiveness and superiority of our proposed method.\\
\majrev{\textbf{Limitations and future works.}
While our Jo-SNC method shows promising results in combating label noise, its performance may degrade on long-tailed datasets with noisy labels. In such scenarios, noisy and tail-class samples can become entangled, compromising the reliability of sample selection and degrading model performance. 
Therefore, our future research will focus on identifying clean samples from long-tailed, noisy datasets, which is a more practical and challenging problem. 
}


\bibliographystyle{bib/IEEEtran}
\bibliography{bib/IEEEabrv,bib/IEEEreference}






\end{document}